\documentclass{vldb}
\usepackage{graphicx}
\usepackage{hyperref}
\usepackage{epstopdf}
\usepackage{balance}  
\usepackage{amsfonts}
\usepackage{url}
\usepackage{subfigure}
\usepackage{algorithm}
\usepackage{amssymb,amsmath}
\usepackage{bbm}
\usepackage{multirow}
\usepackage{ntheorem}   
\usepackage{mdframed}   
\newtheorem{ques}{Question}
\usepackage{color}
\usepackage{ragged2e}
\DeclareMathOperator*{\argmin}{arg\,min}

\DeclareMathOperator*{\avg}{avg}
\newtheorem{theorem}{Theorem}

\newtheorem{problem} {Problem}
\newtheorem{example}{Example}
\newtheorem{lemma}{Lemma}
\definecolor{light-gray}{gray}{0.95}
\def\done{\hspace*{\fill} {{\small $\blacksquare$}}}

\newmdtheoremenv[%
innerleftmargin=0.4pt,%
innerrightmargin=0.4pt,backgroundcolor=light-gray,
innertopmargin=0.4pt,%
innerbottommargin=0.4pt,%
splittopskip=0.4pt,skipbelow=0.4pt,%
skipabove=0.4pt,ntheorem]{myans}{Result}

\begin{document}

\title{Label Propagation on K-partite Graphs with Heterophily}

\numberofauthors{6}
\author{
    \alignauthor
     Dingxiong Deng\\
	       \affaddr{Dept. of Computer Science}\\
	       \affaddr{Univ. of Southern California}\\
	       \email{dingxiod@usc.edu}
	\alignauthor
			Fan Bai\\
	       \affaddr{School of Computer Science}\\
	       \affaddr{Fudan University}
	\alignauthor
			Yiqi Tang\\
	       \affaddr{School of Computer Science}\\
	       \affaddr{Fudan University}
   \and
	\alignauthor
			Shuigeng Zhou\\
	       \affaddr{School of Computer Science}\\
	       \affaddr{Fudan University}\\
	       \email{sgzhou@fudan.edu.cn}
	\alignauthor
			Cyrus Shahabi\\
	       \affaddr{Dept. of Computer Science}\\
	       \affaddr{Univ. of Southern California}\\
	       \email{shahabi@usc.edu}
	\alignauthor
			Linhong Zhu\\
	       \affaddr{Information Sciences Institute}\\
	       \affaddr{Univ. of Southern California}\\
	       \email{linhong@isi.edu}
	}
\maketitle

\begin{abstract}
In this paper, for the first time, we study label propagation in heterogeneous graphs under heterophily assumption. Homophily label propagation (i.e., two connected nodes share similar labels) in homogeneous graph (with same types of vertices and relations) has been extensively studied before. Unfortunately, real-life networks are heterogeneous, they contain different types of vertices (e.g., users, images, texts) and relations (e.g., friendships, co-tagging) and allow for each node to propagate both the same and opposite copy of labels to its neighbors. We propose a $\mathcal{K}$-partite label propagation model to handle the mystifying combination of heterogeneous nodes/relations and heterophily propagation. With this model, we develop a novel label inference algorithm framework with update rules in near-linear time complexity. Since real networks change over time, we devise an incremental approach, which supports fast updates for both new data and evidence (e.g., ground truth labels) with guaranteed efficiency. We further provide a utility function to automatically determine whether an incremental or a re-modeling approach is favored. Extensive experiments on real datasets have verified the effectiveness and efficiency of our approach, and its superiority over the state-of-the-art label propagation methods.
\end{abstract}

\section{Introduction}
Label propagation~\cite{Zhu03semi-supervisedlearning} is one of the classic algorithms to learn the label information for each vertex in a network (or graph). It is a process that each vertex receives labels from neighbors in parallel, then updates its labels and finally sends new labels back to its neighbors. Recently, label propagation has received renewed interests from both academia and industry due to its various applications in many domains such as in spam detection~\cite{abernethy2010graph}, fraud detection~\cite{Faloutsos:2014:LGM:2566486.2576889}, sentiment analysis~\cite{goldberg2006seeing}, and graph partitioning~\cite{Ugander:2013:BLP:2433396.2433461}. Different algorithms~\cite{felzenszwalb2006efficient,ihler2005loopy,SubramanyaJMLR2011,yedidia2003understanding,ZhouBLWS03,Zhu03semi-supervisedlearning} have been proposed to perform label propagation on trees or arbitrary graphs. All of these traditional algorithms simply assume that all vertices are of the same type and restricted to only a single pairwise similarity matrix (graph).

\begin{figure}[!t]
\centering
\includegraphics[width=0.7\columnwidth]{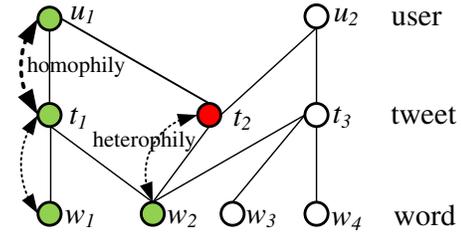}
\caption{{An example of sentiment labeling on a heterogeneous network with three types of vertices: users, tweets and words. Solid edges denote observed links, and dashed edges represent label propagation directions. The width of dash edges distinguishes strengths of propagation. Green color: positive label; red color: negative label; homophily: propagate the same copy of labels; and heterophily: propagate the opposite copy of labels.}}\label{fig:sentiexample}
\vspace{-0.3cm}
\end{figure}

Unfortunately, many real networks such as social networks are heterogeneous systems~\cite{amer2002logical,ShiLZSY15,YangH15} that contain objects of multiple types and are interlinked via various relations. Considering the heterogeneity of data, it is tremendously challenging to analyze and understand such networks through label propagation. Applying traditional label propagation approaches directly to heterogeneous graphs is not feasible due to the following reasons. First, traditional approaches neither support label propagation among different types of vertices, nor distinguish propagation strengths among various types of relations. Consider the example of sentiment labeling shown in Fig.~\ref{fig:sentiexample}, the label of each tweet is estimated using the labels of words and users. In addition, the label information of users is much more reliable than that of words in terms of deciding the labels of tweets~\cite{ZhuGCL14}, and thus the user vertices should have stronger propagation strengths than word vertices. Second, traditional approaches do not support heterophily propagation. As illustrated in Fig.~\ref{fig:sentiexample}, in sentiment label propagation, traditional approaches assume that if a word has a positive label, then its connected tweets also have positive labels. However, it is reasonable that a tweet connected to a positive word is negative due to sarcasm.
There are few notable exceptions~\cite{Ding:2009:LPK:1726586.1727013,GatterbauerVLDB2015,JacobWSDM2014} (see more in related works) that support either heterogeneous types of vertices or heterophily propagation, but not both. Last but not least, all of the current approaches simply assume that the propagation type (e.g., a homophily or heterophily propagation) is given as an input, though actually this is difficult to obtain from observation.

\begin{figure}[!t]
\centering
\subfigure[Folksonomy]{\includegraphics[width=0.46\columnwidth]{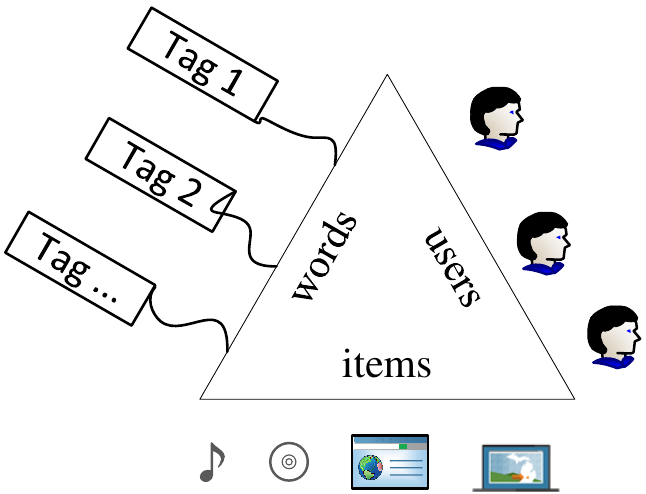}}
\hspace{0.2cm}
\subfigure[Text corpus]{\includegraphics[width=0.44\columnwidth]{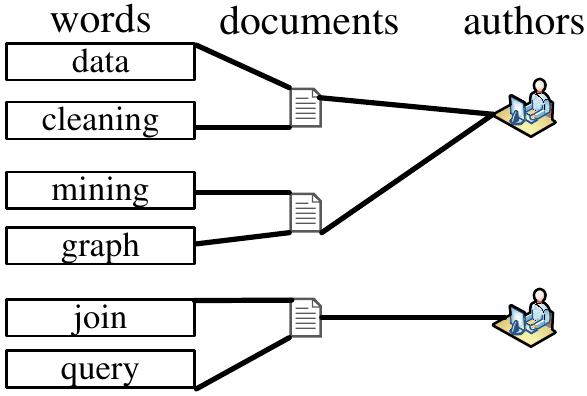}}
\caption{\small{Examples of ubiquitous tripartite graphs.}}\label{fig:realexample}
\vspace{-0.55cm}
\end{figure}

In this paper, we study the problem of labeling nodes with categories (e.g., topics, communities, sentiment classes) in a partially labeled heterogeneous network. To this end, we first propose a $\mathcal{K}$-partite graph model as the knowledge representation for heterogeneous data. Many real-life data examples, naturally form $\mathcal{K}$-partite with different types of nodes and relations. To provide a few examples, as shown in Fig.~\ref{fig:realexample}, in folksonomy system, the triplet relations among users, items and tags can be represented as a tripartite graph; document-word-author relation can also be modeled as another tripartite graph with three types of vertices. A $\mathcal{K}$-partite graph model nicely captures the combination of vertex-level heterogeneity and heterogeneous relations, which motivates us to focus on label propagation on $\mathcal{K}$-partite graphs. That is, given an observed $\mathcal{K}$-partite and a set of very few seed vertices that have ground-truth labels, our goal is to learn the label information of the large number of the remaining vertices. Even though $\mathcal{K}$-partite graphs are ubiquitous and the problem of label propagation on $\mathcal{K}$-partite graphs has significant impact, this area is much less studied as compared to traditional label propagation in homogeneous graphs and thus various modeling and algorithmic challenges remain unsolved.

To address the modeling challenges, we develop a unified $\mathcal{K}$-partite label propagation model with both vertex-level heterogeneity and propagation-level heterogeneity. Consider the sentiment example in Fig.~\ref{fig:sentiexample}, our model allows a tweet vertex to receive labels from both words and users, but automatically gives higher weights to the latter through our propagation matrix. The propagation-level heterogeneity, which is reflected by supporting homophily, heterophily and mixed propagation, are automatically learned in our model. To infer our model, we first propose a framework that supports both multiplicative~\cite{LeeNIPS2000} and addictive rules~\cite{HothesisNMF} (i.e., projected gradient descent) under the vertex-centric manner with near linear-time complexity. Because in practice graph and labels can continuously changing, we then study how and when we should apply label propagation to handle the changing scenarios. We thus devise a fast increment algorithm (i.e., assigning labels for new day, or updating labels upon feedbacks), that performs partial updates instead of re-running the label inference from scratch. We not only can control the trade-off between efficiency and accuracy via the confidence level parameter for speed up, but also can determine whether we should apply our incremental algorithm through a utility function.

The contributions of this work are as follows:

\begin{enumerate}
{
\item \textbf{Problem formulation}: we show how real-life data analytic tasks can be formulated as the problem of label propagation on $\mathcal{K}$-partite graphs.
\item \textbf{Heterogeneity and heterophily}: we propose a generic label propagation model that supports both heterogeneity and heterophily propagation.
\item \textbf{Fast inference algorithms}: we develop a unified label propagation framework on $\mathcal{K}$-partite graphs that supports both multiplicative and addictive rules with \emph{near-linear time complexity}. We then propose an incremental framework, which supports much faster updates upon new data and labels than re-computing from scratch. In order to strike a balance between efficiency and effectiveness, we introduce a confidence level parameter for speed up. We further develop a utility function is designed to automatically choose between an incremental or a re-computing approach.
\item \textbf{Practical applications}: we demonstrate with three typical application examples that various classification tasks in real scenarios can be solved by our label propagation framework with $\mathcal{K}$-partite graphs.
    }
\end{enumerate}


\section{Related Works}
\begin{table}[!t]
  \centering
\caption{A summary for label propagation methods. H-V denotes whether a method supports heterogeneous types of vertices, H-P denotes whether a method supports heterophily in label propagation, Auto means whether the heterophily  matrix is automatically learned or predefined, Incre denotes whether the proposed label propagation algorithm supports incremental update upon new data or new labels, and Joint denotes whether the proposed approach allows a vertex to be associated with multiple labels.}\label{tab:related}
  \begin{tabular}{|l|l|l|l|l|l|}
    \hline
    Method&H-V&H-P&Auto&Incre&Joint\\
    \hline
    ~\cite{Blum:2004:SLU:1015330.1015429}~\cite{felzenszwalb2006efficient}~\cite{SubramanyaJMLR2011}~\cite{talukdar2009new}
&\multirow{2}{*}{X}
&\multirow{2}{*}{X}
&\multirow{2}{*}{X}
&\multirow{2}{*}{X}
&\multirow{2}{*}{X}\\
~\cite{yedidia2003understanding}~\cite{ZhouBLWS03}~\cite{Zhu03semi-supervisedlearning}
&
&
&
&
&\\
\hline
 ~\cite{ChakrabartiFCM14}&X&X&X&X&$\checkmark$\\
    \hline
    ~\cite{Ding:2009:LPK:1726586.1727013}~\cite{JacobWSDM2014}&$\checkmark$&X&X&X&?\\
    \hline
    ~\cite{GatterbauerVLDB2015}~\cite{Yamaguchi:2015:OSN:2888116.2888151}&X&$\checkmark$&X&$\checkmark$&X\\
    \hline
    Proposed&$\checkmark$&$\checkmark$&$\checkmark$&$\checkmark$&$\checkmark$\\
    \hline
  \end{tabular}
  \vspace{-0.2cm}
\end{table}
In this section, we first provide an extensive (but not exhaustive) review about the state-of-the-art label propagation approaches, and then discuss related works on heterogeneous graph representation using $\mathcal{K}$-partite graphs.

\subsection{Label Propagation}
We summarize a set of representative label propagation algorithms in Table~\ref{tab:related}. In the following, we provide more details about each approach, with emphasis on their advantages and disadvantages. First we consider how these methods support propagating labels in heterogeneous networks. Various types of algorithms have been proposed to perform label propagation on trees or arbitrary graphs such as belief propagation~\cite{felzenszwalb2006efficient}~\cite{yedidia2003understanding}, loopy belief propagation~\cite{ihler2005loopy}, Gaussian Random Field (GRF)~\cite{Zhu03semi-supervisedlearning}, MP~\cite{SubramanyaJMLR2011}, MAD~\cite{talukdar2009new}, and local consistency~\cite{ZhouBLWS03}. All of these algorithms simply assume that all vertices are of the same type and restricted to only a single graph.

Recently with the tremendously increasing of heterogeneous data, label propagation approaches on heterogeneous networks have been developed. Jacob et al.~\cite{JacobWSDM2014} focused on learning the unified latent space representation through supervised learning for heterogeneous networks. Ding et al.~\cite{Ding:2009:LPK:1726586.1727013} proposed a cross propagation algorithm for $\mathcal{K}$-partite graphs, which distinguishes vertices of different types, and propagates label information from vertices of one type to vertices of another type. However, they still make the homophily propagation assumption, that is, two connected nodes have similar labels (representations), even though they are from different types. On another hand, Gatterbauer et al.~\cite{GatterbauerVLDB2015} proposed a heterophily belief propagation algorithm for homogeneous graphs with the same type of vertices (e.g., a user graph). Yamaguchi et al.~\cite{Yamaguchi:2015:OSN:2888116.2888151} also proposed a heterophily propagation algorithm that connects to random walk and Gaussian Random Field. In both approaches, they assumed that even a user node is labeled as fraud, it can not simply propagate the fraud label to all its neighbors. However, their approach does not support vertex-level heterogeneity and use the same type of propagation over the entire network. In addition, the propagation matrices (e.g., diagonal matrix for homophily, off-diagonal matrix for heterophily) are required to be predefined based on observation, rather than to be automatically learned. In this work, we propose a unified label inference framework, which supports both vertex-level heterogeneity, and propagation-level heterophily (i.e., different types of propagation across heterogeneous relations). Furthermore, our framework is able to automatically learn the propagation matrices from the observed networks.

Current researches have been focused on addressing algorithmic issues in the problem of label propagation such as incrementally and jointly learning for label inference.  Gatterbauer et al.~\cite{GatterbauerVLDB2015} proposed a heuristic incremental belief propagation algorithm with new data. However, more research is required in finding provable performance guarantee for the incremental algorithm. Chakrabarti et al.~\cite{ChakrabartiFCM14} proposed a framework with joint inference of label types such as hometown, current city, and employers, for users connected in a social network. To advance existing work, we propose an incremental framework which supports adaptive update upon both new data and labels. In addition, our algorithm is guaranteed to achieve speedup compared to re-computing algorithms with a certain confidence. It also supports multi-class label joint inference and provides a better understanding about the latent factors that cause link formations in the observed heterogeneous networks.

\subsection{$\mathcal{K}$-partite Graphs}
$\mathcal{K}$-partite graph analysis has wide applications in many domains such as topic modeling~\cite{Long:2006:ULK:1150402.1150439}, community detection~\cite{PeiCS15}, and sentiment analysis~\cite{ZhuGCL14}. Most of these works use tripartite graph modeling as a unified knowledge representation for heterogeneous data, and then formulate the real data analytic tasks as the corresponding tripartite graph clustering problem. For example, Long et al.~\cite{Long:2006:ULK:1150402.1150439} proposed a general model, the relation summary network, to find the hidden structures (the local cluster structures and the global community structures) from a $\mathcal{K}$-partite graph; Zhu et al.~\cite{ZhuGCL14} addressed both static tripartite graph clustering and online tripartite graph clustering with matrices co-factorization. There are other works which study theoretical issues such as competition numbers of tripartite graphs~\cite{kim2008competition}.

To summarize, $\mathcal{K}$-partite graph modeling and analysis has been studied from different perspectives due to its potential in various important applications. Yet studies on learning with $\mathcal{K}$-partite graph modeling are limited. In this work, we formulate a set of traditional classification tasks such as sentiment classification, topic categorization, and rating prediction as the label propagation problem on $\mathcal{K}$-partite graphs. With the observed $\mathcal{K}$-partite graphs, our label inference approach is able to obtain decent accuracy with very few labels.

\vspace{-3mm}
\section{Problem Formulation}
Let $t, t'$ be a specific vertex type. A $\mathcal{K}$-partite graph $G=$ $<\cup_{t=1}^{\mathcal{K}}V_t$, $\cup_{t=1}^{\mathcal{K}}\cup_{t'=1}^{\mathcal{K}}E_{tt'}>$ has $\mathcal{K}$ types of vertices, and contains at most $\mathcal{K} (\mathcal{K}-1)/2$ two-way relations. We also use notation $G$ to denote the adjacency matrix representation of a $\mathcal{K}$-partite graph and use $G_{tt'}$ to denote the sub graph/matrix induced by the set of $t$-type vertices $V_t$ and the set of $t'$-type vertices $V_{t'}$. For ease of presentation, Table~\ref{tab:notations} lists the notations we use throughout this paper.

Intuitively, the purpose of label propagation on $\mathcal{K}$-partite graph is to utilize observed labels of seed vertices (denoted as $V^L$) to infer label information for remaining unlabeled vertices (denoted as $V^U$). For each vertex $v$, we use a column vector $Y(v)\in R^{k\times 1}$ to denote the probabilistic label assignment to vertex $v$ and a matrix $Y\in R^{n\times k}$ to denote the probabilistic label assignment for all the vertices, where $n$ is number of nodes and $k$ is number of labels. Therefore, our objective is to infer the $Y$ matrix so that each unlabeled vertex $v \in V^{U}$ obtains an accurate estimation of its ground truth label. 

\begin{figure}[!t]
	\centering
	\includegraphics[width=\columnwidth]{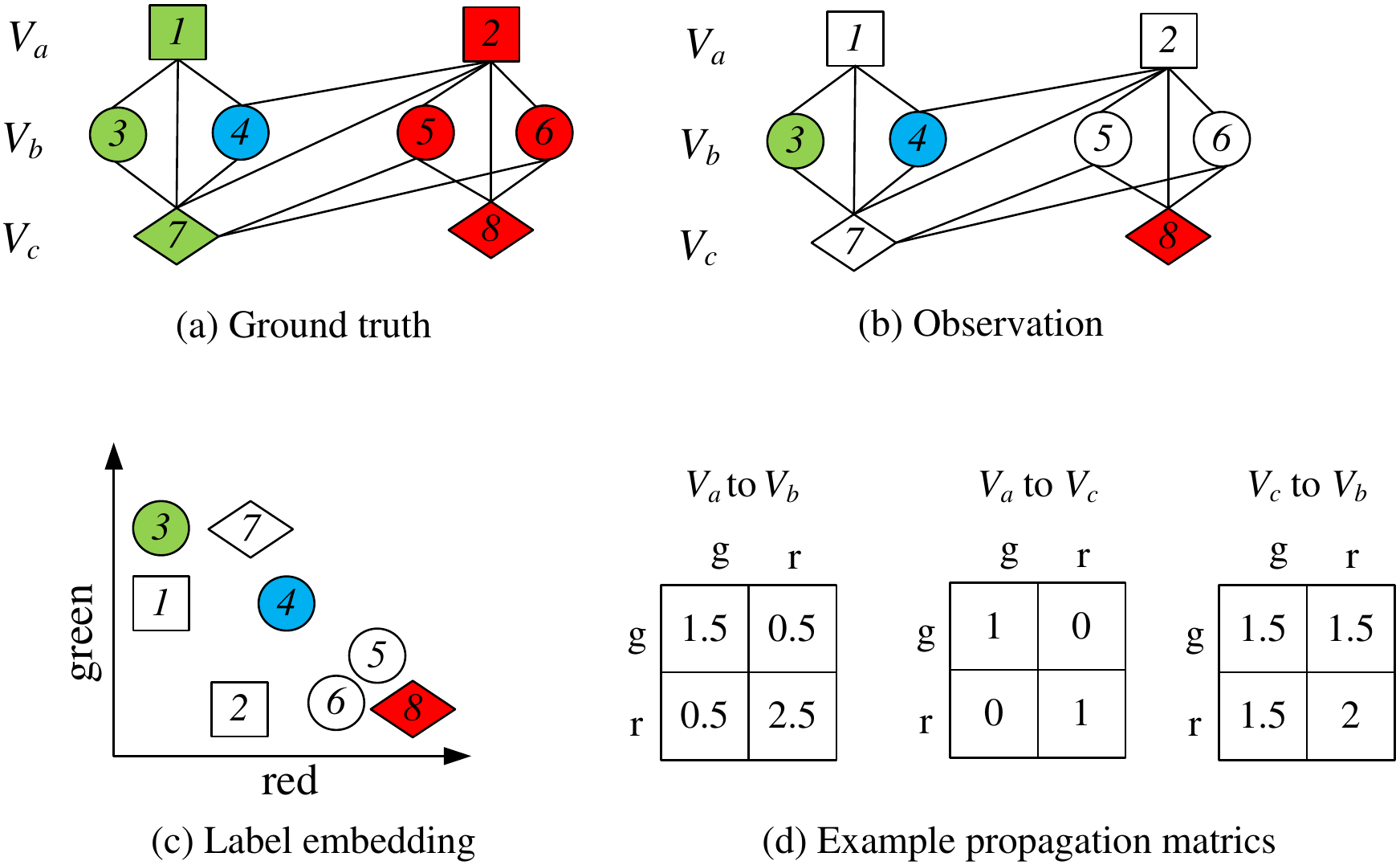}
	\caption{\small{An illustration of our label inference process with a tripartite graph example. Here we use different shapes/colors to distinguish types/labels. g: green, and r: red. In addition, the blue color denotes an overlapping assignment of both green and red.}}\label{fig:tripartiteframework}
\end{figure}

We first build intuition about our model using a running example shown in Fig.~\ref{fig:tripartiteframework}(a). Suppose we only observe the links and very few labels of vertices as shown in Fig.~\ref{fig:tripartiteframework}(b), the purpose of label propagation on tripartite graphs is then to utilize observed links to infer label information. We introduce a matrix $B\in R^{k\times k}$ to represent the correlation between link formation and associated labels. Each entry $B(l_i, l_j)$ denotes the likelihood that a node labeled as $l_i$ is connected with another node labeled as $l_j$. Subsequently, in the label propagation process, each entry $B(l_i, l_j)$ also denotes a proportional propagation that indicates the relative influence of nodes with label $l_i$ on another nodes with label $l_j$. To keep consistent with the label propagation process, here we interpret $B$ as the propagation matrix.

Note that the $B$ matrix is an arbitrary non-negative matrix without any restrictions. If the $B$ matrix is diagonal (e.g., $B_{ac}$ from $V_a$ to $V_c$ in Fig.~\ref{fig:tripartiteframework}(d)), the link formation is consistent with homophily assumption; while if the $B$ matrix is off-diagonal, the link formation is more likely due to heterophily assumption. In addition, as shown in Fig.~\ref{fig:tripartiteframework}(d), the propagation matrix between $a-$ and $b-$ type vertices is different from that between $a-$ and $c-$ type vertices. We thus let $B_{tt'}$ denote the propagation matrix between $t$-type vertices and $t'$-type vertices.

Generally speaking, if we have prior knowledge about propagation matrix $B$, then each unlabeled vertex could receive proportional propagation from very few labeled vertices according to $B$. We propose to infer the propagation matrix and label assignment via embedding. We embed each vertex into a unified space (for all types of vertices), where each dimension of the latent space denotes a specific label and the link probability of two vertices are correlated to their labels (i.e., distance in the latent space). Fig.~\ref{fig:tripartiteframework}(c) shows a simplified two-dimension space. With the label embedding, the label assignment and the correlation can be automatically learned.

With these intuitions, we focus on the following problem:
\vspace{-5mm}
\begin{problem}\emph{(\textbf{$\mathcal{K}$-partite Graph Label Inference})}\label{problem:all}
	Given a $\mathcal{K}$-partite graph, a set of labels $L$, a set of seed labeled vertices $V^L$ with ground truth labels $Y^*\in R^{n_L\times k}$ $(n_L=|V^L|)$, the goal is to infer the label assignment matrix $Y$ and the propagation matrix $B$ such that
\begin{equation}\label{eq:obj}
{
		\begin{aligned}
			&\arg\min_{Y, B\geq 0}\{\sum_{t}\sum_{t^{\prime}\neq t}\|G_{tt^{\prime}}-Y_{t}B_{tt^{\prime}}Y_{t^{\prime}}^T\|_F^2\\
			&+\beta\sum_{u\in V^L}\|Y(u)-Y^*(u)\|_F^2+\lambda\emph{\texttt{regularizer}}(G, Y)\}
		\end{aligned}
}
\end{equation}
where $t$ denotes the type of vertex, $Y_t$ denote a sub matrix of $Y$, which gives the label assignment for the set of $t$-type vertices $V_t$, 
$\beta$ and $\lambda$ are parameters that control the contribution of different terms, and $\texttt{regularizer}(G, Y)$ denotes a regularization approach such as graph regularization~\cite{CaiPAMI11}, sparsity~\cite{GilpinKDD2013}, diversity~\cite{yu2009recommendation}, and complexity regularization.
\end{problem}

In our objective function, the first term evaluates how well each $B$ matrix represents the correlation between link formation $G$ and associated labels $Y$, via computing the error between estimated link formation probability and the observed graph. The second term $\sum_{u}\|Y(u)-Y^*(u)\|_F^2$ gives penalty to seed vertices if their learned labels are far away from the ground truths. Note that the regularization term provides an add-on property that utilizes additional domain-specific knowledge to improve learning accuracy.


In conclusion, our problem definition has well addressed all the mentioned modeling challenges. Unfortunately, besides modeling challenges, there are several computational challenges remaining to be solved. First, the NP-hardness of Problem~\ref{problem:all} (the sub problem of nonnegative matrix factorization is NP-hard~\cite{Vavasis2009}) requires efficient solutions for large-scale real problems. Second, the rapid growth of data and feedbacks requires fast incremental update. In the following, we address those computational challenges by developing efficient algorithms that are highly scalable and achieve high quality in terms of classification accuracy for real-life tasks. 

\begin{table}[!t]
	\small{
		\centering
		\caption{Notations and explanations.}\label{tab:notations}
		\vspace{-0.5mm}
		{
			\begin{tabular}{|l|l|}
				\hline
				Notations & Explanations. \\
				\hline
				$n$/$m$/$k$/$\mathcal{K}$& number of nodes/edges/labels/types\\
				\hline
				$G$/$B$/$Y$& Graph/Propagation/Label assignment matrix\\
				\hline
				$t(v)/d(v)/N(v)$& the type/degree/neighbors of vertex $v$\\
				\hline
				$Y(v)\in R^{k\times 1}$& the label assignment for vertex $v$\\
				\hline
				\multirow{2}{*}{$M_{tt'}$}&a sub matrix of matrix $M$ between t-type \\
				&vertices and t'-type vertices \\
				\hline
				$\eta$ & step size in additive rule\\
				\hline
				$\epsilon$ & a very small positive constant\\
				\hline
				$\theta$& parameter in incremental algorithm\\
				\hline
				$1_A(x)$ &indicator function\\
				\hline
				$\mathcal{L}$&Lipschitz constant\\
				\hline
				$J(x)$&Objective function of $x$\\
				\hline
				$\mathcal{J}(x)$&Largrangian function of $x$\\
				\hline
			\end{tabular}
		}}
		\vspace{-0.0cm}
	\end{table}

\section{Label Inference Algorithms}
In this section, we discuss the algorithmic issues in label inference process. Specifically, our label inference problem leads to an optimization problem, which searches for the optimum weighted propagation matrix $B$ and label assignment matrix $Y$ to minimize the objective function in Eq.~(\ref{eq:obj}). Considering these nonnegative constraints, different approaches~\cite{HothesisNMF,LinNC2007, Kim:2014:ANM:2582309.2582329} have been proposed to solve this non-convex optimization problem. Among them, the multiplicative update~\cite{LeeNIPS2000} and additive update rules~\cite{LinNC2007} are two most popular approaches because of their effectiveness. The multiplicative update approach batch updates each entry of the matrix by multiplying a positive coefficient at each iteration, while the additive update approach is a project gradient descent method. To the best of our knowledge, there is no existing work that combines different update rules in a unified framework.

In this paper, we propose to combine these two update rules in a unified framework. We observe that the typical multiplicative update rule that batch updates each entry of the matrix, can be transferred to a vertex-centric rule that corresponds to update each row of the matrix per iteration. This transformation allows us to unify both rules under the same vertex-centric label propagation framework because many addictive rules are updating each row per iterations~\cite{LinNC2007, Kim:2014:ANM:2582309.2582329}. We notice that multiplicative and addictive update rules share some common computations. Consequently by pre-computing these common terms, our framework can be independent to various different update rules, while remains as efficient as possible. The proposed framework enjoys three important by-products: (1) supporting fair comparison between different update rules, (2) one unified incremental algorithm in Section~\ref{subsec:incre} naturally support various update rules, (3) easy to parallel because the updates of each vertex can be performed at the same time.


\begin{algorithm}[!t]
	\caption{The unified label inference framework}\label{alg:framework}
		\begin{tabbing}
			\textbf{Input}: Graph $G$, a few ground truths $Y^*$\\
			\textbf{Output}: Label Matrix $Y$\\
			01: Initialize $Y$ and $B$ (see Section~\ref{subsec:Init})\\
			02: \textbf{repeat}\\
			03: \hspace{0.5cm}update $B_{tt^{\prime}}$ (see Section~\ref{subsec:B})\\
			04: \hspace{0.5cm}update common terms $A_t$ (see Eq.~(\ref{equ:m:A}))\\
			\hspace{1.0cm}$/*$ Vertex-centric search (block coordinate search)$*/$\\
			05: \hspace{0.5cm}\textbf{for} each vertex $u\in V$\\
			06: \hspace{1.0cm}update and/or normalize $Y(u)$ (see Section~\ref{subsec:Y})\\
			07: \hspace{0.5cm}$Y_s=Y$; \\
			08: \hspace{0.5cm}{\small $Y=\argmin_{Y}||Y_s-Y||_F^2+\lambda\texttt{regularizer}(G, Y)$}\\
			09: \textbf{until} converges\\
			10: \textbf{return} $Y$
		\end{tabbing}
	\vspace{-0.4cm}
\end{algorithm}

Algorithm~\ref{alg:framework} presents the framework of our vertex-centric label inference algorithm, which consists of three steps: initialization, update for the propagation matrix $B$, and update for the label assignment of each vertex $Y(u)$. We first provide a non-negative initialization for both $Y$ and $B$ in Section~\ref{subsec:Init}. We next iteratively update $B$ (Section~\ref{subsec:B}) and $Y$ (Section~\ref{subsec:Y}) until the solution converges (Lines 2--10) with both multiplicative and addictive rules. Because the computational cost is dominated by updating the label assignments for all the vertices (Lines 5--6). To reduce the computational cost,  when updating label assignment matrix $Y$ in Section~\ref{subsec:Y}, we design an efficient cache technique that pre-computes and reuses common terms shared by the same type vertices (i.e., $A_t$ for each $t$-type vertex) for both update rules. We show that by pre-computing $A_t$, the computational time for each $Y_t$ decreases, consequently the computational cost per iteration is much reduced.
 
In the following we first present the details of three components in our unified algorithm framework from Section~\ref{subsec:Init}--Section~\ref{subsec:B} , we then show the equivalence between element-wise multiplicative and vertex-centric multiplicative, and analyze the theoretical properties of different update rules in Section~\ref{subsec:comp}.

\subsection{Initialization}\label{subsec:Init}
To achieve a better local optimum, a label inference algorithm should start from one or more relatively good initial guesses. In this work, we focus on graph proximity based initialization for label assignment matrix $Y$; while $B$ matrix is initialized using observed label propagation information among labeled seed vertices.

\vspace{0.1cm}
\noindent \textbf{Initializing $\mathbf{Y}$} Given the set of labeled vertices $V^L$ with ground truth label $Y^*\in R^{n_l\times k}$, we utilize the graph proximity to initialize the unlabeled vertices with similar labeled and the same-type vertices. Specifically, the label assignment matrix $Y^0$ can be initialized as follows:
\begin{equation}\label{equ:init2}
\small{
Y^0(u) = \begin{cases} Y^*(u) &\mbox{if } u \in V^L \\
\avg\limits_{v\in V_{t(u)}}\texttt{sim}(u, v, G)Y(v)&\mbox{otherwise}\end{cases}
}
\end{equation}
where $\texttt{sim}(u, v)$ evaluates the graph proximity between vertices $u$ and $v$. In our experiments, we define $\texttt{sim}(u, v, G)$ as the normalized Admic-Adar score~\cite{Adamic01friendsand}. In order to evaluate the similarity between two vertices, we sum the number of neighbors the two vertices have in
common. Neighbors that are unique to a few vertices are weighted more than commonly occurring neighbors. That is,
\begin{equation}\label{equ:simgraph}
\texttt{sim}(u, v, G)=\sum_{w\in (N(u)\cap N(v))}\frac{1}{\log d(w)}
\end{equation}
We first compute the $\texttt{sim}(u, v, G)$ with Eq.~(\ref{equ:simgraph}), and then normalize all the scores into the range $[0, 1]$. Here $d(u)$ is the degree of vertex $u$, and $N(u)$ is the set of neighbors of vertex $u$.

\vspace{0.1cm}
\noindent \textbf{Initializing $\mathbf{B}$} For each $B_{tt'}$, we initialize it based on label class information and vertex type information of the observed labeled vertices. Specifically, we first initialize each $B_{tt'}$ as an identity matrix, where we assume that if type-$t$ vertices are in $l_i$ class, then all their connected type-$t'$ vertices will receive corresponding $l_i$ class label. We then increment $B_{tt'} (l_i, l_j)$ by one whenever we observe a type-$t$ vertex labeled $l_i$ is connected to another type-$t'$ vertex labeled $l_j$. Finally, we normalize each $B_{tt'}$ using $L_1$ norm.

\subsection{Update $\mathbf{Y}$}\label{subsec:Y}
As introduced earlier, we perform vertex-centric update for $Y$. That is, in each iteration, we focus on minimizing the following sub objective:
\begin{equation}
\small{
\begin{aligned}\label{equ:subobj}
J(Y(u))&=\sum_{v\in N(u)}(G(u,v)-Y(u)^TB_{t(u)t(v)}Y(v))^2\\
&+\sum_{v\not\in N(u), t(v)\neq t(u)}(Y(u)^TB_{t(u)t(v)}Y(v))^2\\
&+\mathbf{1}_{V^L}(u)\beta\|Y(u)-Y^*(u)\|_F^2
\end{aligned}
}
\end{equation}
where $1_A(x)$ is the indicator function, which is one if $x\in A$ and zero otherwise, $N(u)$ is the neighbors of vertex $u$.
%

In the following, we adopt two representative update rules, multiplicative rule and additive rule, to derive the optimal solution for Eq.~(\ref{equ:subobj}).

\begin{lemma}\label{lemma:MAY}{(\emph{\textbf{Multiplicative rule for $\mathbf{Y}$}})
$Y$ can be approximated via the following multiplicative update rule:
\begin{equation}\label{equ:m:Y}
{
\begin{aligned}
&Y(u)=Y(u)\circ\\
&\sqrt{\frac{\sum_{v\in N(u)}G(u,v)B_{t(u)t(v)}Y(v)+\beta\mathbf{1}_{V^L}(u)Y^{*}(u)}{A_{t(u)}Y(u)+\beta\mathbf{1}_{V^L}(u)Y(u)+\epsilon}}
\end{aligned}
}
\end{equation}
where $\epsilon>0$ is a very small positive value (e.g., $1^{-9}$), and $A_t$ is defined as follows:
\begin{equation}\label{equ:m:A}
\small{
A_t=\sum_{v\not\in V_{t}}B_{tt(v)}Y(v)Y(v)^TB_{tt(v)}^T
}
\end{equation}
}
\end{lemma}
\noindent \textbf{Proof}: The proof can be derived in spirit of the classic multiplicative algorithm~\cite{LeeNIPS2000} for Non-negative matrix factorization with the KKT condition~\cite{kuhn50nonlinear}. Details are presented in Appendix~\ref{proof:2}.

\begin{lemma}(\emph{\textbf{Additive rule for $\mathbf{Y}$}})
An alternative approximate solution to $Y$ can be derived via the following additive rule:
\begin{equation}\label{equ:A:Y}
{
\begin{aligned}
Y(u)^{r+1}=&\max(\epsilon,Y(u)^r+2\eta(\sum_{v\in N(u)}G(u,v)B_{t(u)t(v)}Y(v)\\
&-A_tY(u)+\beta\mathbf{1}_{V^L}(u)(Y^{*}(u)-Y(u)^{r})))
\end{aligned}
}
\end{equation}
where $\eta$ is the step size, and $A_t$ is defined in Eq.~(\ref{equ:m:A}).
\end{lemma}
\noindent \textbf{Proof}: It can be easily derived by replacing the deviation of $J(Y(u))$ into the standard gradient descent update rule.

\vspace{0.1cm}
\noindent \textbf{Step size for additive rule.} We use Nesterov's method~\cite{GuanTLY12},\cite{Nesterov04} and \cite{ZhuTKDE16} to compute the step size $\eta$, which can be estimated using the Lipschitz constant $\mathcal{L}$ for $\nabla J(Y(u))$, see Appendix~\ref{proof:constant}.

\subsection{Update $\mathbf{B}$}\label{subsec:B}
In the following, we present the detailed update rules for propagation B.

\begin{lemma}\label{lemma:MAB}(\textbf{\emph{Multiplicative rule for $\textbf{B}$}})
$B$ can be derived via the following update rule:
\begin{equation}\label{equ:m:Bab}
{
B_{tt^{\prime}}=B_{tt^{\prime}}\circ\sqrt{\frac{Y_t^TG_{tt^{\prime}}Y_{t^{\prime}}}{Y_t^TY_tB_{tt^{\prime}}Y_{t^{\prime}}^TY_{t^{\prime}}+\epsilon}}
}
\end{equation}
\end{lemma}
\noindent \textbf{Proof}: Proof of this Lemma is similar to Lemma~\ref{lemma:MAY}, as shown in Appendix~\ref{proof:4}.

\begin{lemma}(\textbf{\emph{Additive rule for $\textbf{B}$}})
An alternative approximate solution to $B$ can be derived via the following additive rule:
\begin{equation}\label{equ:A:Bab}
{
B_{tt^{\prime}}=\max(\epsilon,B_{tt^{\prime}}+2\eta_b(Y_t^TG_{tt^{\prime}}Y_{t^{\prime}}-Y_t^TY_tB_{tt^{\prime}}Y_{t^{\prime}}^TY_{t^{\prime}}))
}
\end{equation}
where $\eta_b$ again denotes the step size.
\end{lemma}
\noindent \textbf{Proof}: It can be easily derived by replacing the deviation of $J(B_{tt^{\prime}})$ into the standard gradient descent update rule.

Similar to the computation of $\eta$ with Nesterov's gradient method, $\eta_b$ can be computed with Lipschitz constant $\mathcal{L}_b$$=2\|Y_{t'}^TY_{t'}Y_t^TY_t\|_F$ for $\nabla J(B_{tt^{\prime}})$, see Appendix~\ref{proof:constant2}.

\subsection{Comparison and Analysis}\label{subsec:comp}
\begin{table}[!t]
	\centering
	\caption{Time complexity of basic operators, where $n$ is number of nodes, $m$ is number of edges, and $k$ is number of label classes.}\label{tab:complexity}
	\begin{tabular}{|c|c|c|c|}
		\hline
		&$B$ & $Y$&$A_t$\\
		\hline
		Multi&$O(n+m)k$  & $O(n+m)k$ &$\sum_{t'\neq t}n_{t'}k^2$\\
		\hline
		Addti&$O(n+m)k$  & $O(n+m)k$ &$\sum_{t'\neq t}n_{t'}k^2$\\
		\hline
	\end{tabular}
\end{table}

We first show that the solution $Y$ returned by the proposed multiplicative rule is identical to that by the traditional multiplicative rule proved in the following lemma.

\begin{lemma}\label{lemma:identical}
	Updating label assignment $Y$ vertex by vertex using Lemma~\ref{lemma:MAY} is identical to the following traditional multiplicative rule~\cite{ZhuGCL14}:
	\begin{equation*}
	Y_{t}=Y_t\circ\sqrt{\frac{\sum_{t^{\prime}\neq t}G_{tt^{\prime}}Y_tB_{tt^{\prime}}^T+\beta S_tY_0}{\sum_{t^{\prime}\neq t}Y_tB_{tt^{\prime}}Y_{t^{\prime}}^TY_{t^{\prime}}B_{tt^{\prime}}^T+\beta S_tY_t}}
	\end{equation*}
	where $S\in R^{n\times n}$ is the label indicator matrix, of which $S_{uu}=1$ if $u\in V^L$ and zero for all the other entries, and $S_t$ is the sub matrix of $S$ for $t$-type vertices.
\end{lemma}
\noindent \textbf{Proof}: The detailed proof is shown in Appendix~\ref{proof:identical}.

We then analyze the time complexity of computing each basic operator in a single iteration. As outlined in Table~\ref{tab:complexity}, each basic operator can be computed efficiently in real sparse networks. In addition, because of our pre-computation of $A_t$, multiplicative and additive update rules have the same near-linear time computational cost in a single iteration, which is much smaller than many traditional multiplicative rules for $Y$. For example, if we apply the multiplicative rule proposed by Zhu et. al.~\cite{ZhuGCL14}, it leads to $O(n_an_bk+n_an_ck+n_bn_ck)$ computation complexity per iteration.


\noindent \textbf{Convergence.} Let us first examine the general convergence of our label inference framework. Based on Corollary 1, 2 and 3~\cite{Kim:2014:ANM:2582309.2582329}, any limited point of the sequence generated by Algorithm~\ref{alg:framework} reaches the stationary point if the update rules remain non-zero and achieve optimum. Therefore, in general, if any update rule in Algorithm~\ref{alg:framework} is optimum for both sub objectives, it leads to the stationary point.

We next analyze the convergence properties when using both multiplicative update rules and additive update rules. Although both the multiplicative updating rules and additive rules used in this work are not optimum for subproblems, they still have very nice convergence properties. As proved on Theorem 2.4~\cite{Calamai1987}, the proposed additive rules still converge into a stationary point. For the multiplicative rule, we conclude that using the proposed multiplicative updating rules guarantees that the value of objective function is non-increasing, and thus the algorithm converges into a local optima. This is because the proposed multiplicative rules are identical to the traditional multiplicative rules as proved in Lemma~\ref{lemma:identical}, and Zhu et al.~\cite{ZhuGCL14} have proved that the value of objective function is non-increasing with the traditional multiplicative rules.

\section{Incremental Label Inference}\label{subsec:incre}

In this section, we present our solution to the fundamental research question with practical importance: \texttt{How can we support fast incremental updates upon graph updates such as new labels and/or new node/edges?} This is because in practice graphs and labels are continuously changing. In the following, we first develop an incremental algorithm that adaptively updates label assignment upon new data, where \textit{we can control the trade-off between efficiency and accuracy}. We then further explore another interesting question: \texttt{on which condition it is faster to perform incremental update than recomputing from scratch?} To address this issue, we propose a utility function that examines the reward and cost of both update operations. With the utility function, our framework is able to automatically determine the ``best" strategy based on different levels of changes.

\subsection{Incremental Update Algorithm}
Our incremental update algorithm supports both graph changes and label changes. The first scenario includes vertex/edge insertion and deletion; while the label changes include receiving additional labels, or correction of noise labels. With the proliferation of crowdsourcing, whenever we get additional explicit labels from users; or we identify noise labels based on user feedback, we can update label assignment for each vertex.

\begin{algorithm}[!t]
\caption{Incremental label Inference algorithm with changes}\label{alg:Incremdata}
\begin{tabbing}
\textbf{Input}: Graph $G$, old label matrix $Y$, a few ground truths $Y^*$,\\
\hspace{1.05cm} confidence level $\theta\in[0, 1)$, and changes $\Delta V$\\
\textbf{Output}: New label matrix $Y_n$\\
01: $\texttt{cand}$=$\Delta V$\\
02: for each $u\in G$\\
03: \hspace{0.5cm}$Y_n(u)=Y(u)$\\
04:  $w_{tt'}=\texttt{avg}_{(u,v)\in E_{tt'}}Y(u)^TB_{tt'}Y(v)$, \\
\hspace{0.55cm}$\sigma_{tt'}=\texttt{std}_{(u,v)\in E_{tt'}}Y(u)^TB_{tt'}Y(v)$\\
05: \textbf{repeat}\\
06: \hspace{0.5cm}for each vertex $u\in \texttt{cand}$\\
07: \hspace{1.0cm}update $Y_n(u)$ (see Section~\ref{subsec:Y})\\
08: \hspace{0.5cm}for each $v\in N(u), v\not\in \texttt{cand}$\\
09: \hspace{1.0cm}if {\small $|Y_n^{T}(u)B_{t(u)t(v)}Y_n(v)-w_{t(u)t(v)}|\geq$}{\small $\sqrt{\frac{1}{1-\theta}}\sigma_{t(u)t(v)}$}\\
10: \hspace{1.5cm}$\texttt{cand}=\texttt{cand}\cup \{v\}$\\
11: \textbf{until} $Y_n$ converges\\
12: \textbf{return} $Y_n$
\end{tabbing}
\vspace{-4mm}
\end{algorithm}

A  simple approach to deal with graph/label updates is to re-run our label inference algorithm for the entire graph. However, this can be computationally expensive. We thus propose the incremental algorithm, where we perform partial updates instead of full updates for each vertex. Our incremental algorithm is built upon the intuition that even with the graph and label changes, the majority of vertices tend to keep the label assignments or will not change much. Therefore, we perform a "lazy" adjustment, i.e., utilizing the old label assignment and updating a small candidate set of change vertices. Unfortunately, it is very challenging to design an efficient and effective incremental label propagation algorithm because in label propagation the adjustment of existing vertices will affect its neighbors, and even vertices that are far away. This requires us to propose an effective strategy to choose which subset of vertices to be adjusted to guarantee the performance gain of the incremental algorithm. Let us use the following example to illustrate more about the challenge.

\begin{figure}[!thb]
  \centering
  \includegraphics[width=\columnwidth]{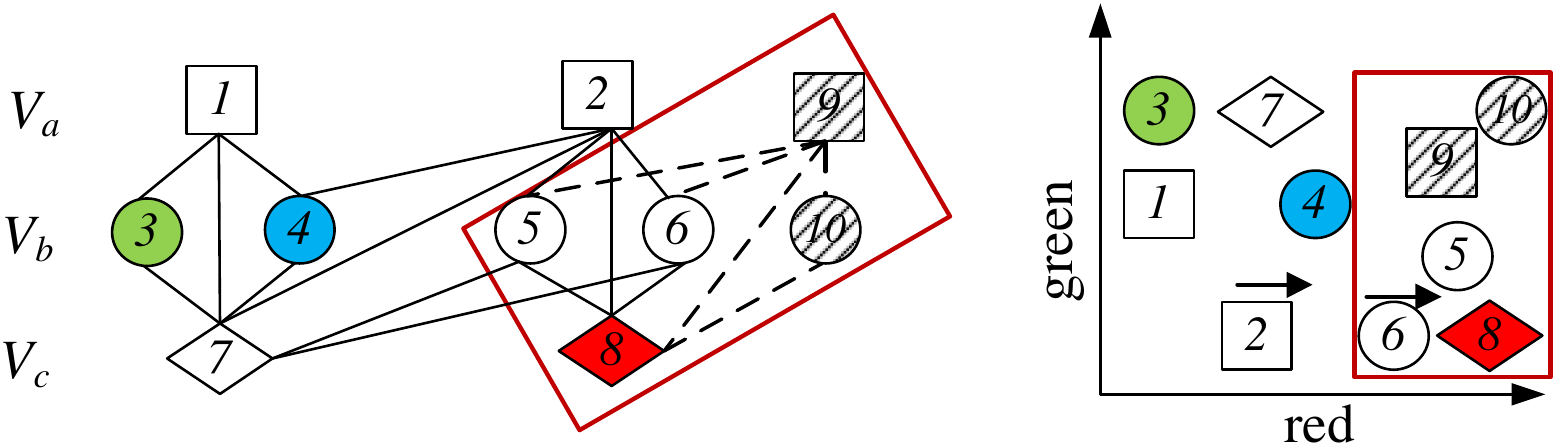}
  \vspace{-5mm}
  \caption{Two new vertices are inserted into the Tripartite graph in Fig.~\ref{fig:tripartiteframework}.}\label{fig:incre1}
    \vspace{-0.0cm}
\end{figure}


\vspace{-1mm}
\begin{example}
Fig.~\ref{fig:incre1} shows an example of initializing the candidate set of vertices to be adjusted when graph receive updates. Two vertices 9 and 10 are inserted with new formed edges into the tripartite graph shown in Fig.~\ref{fig:tripartiteframework} (a), which consequently leads to vertices 5, 6, 8 receive new links from vertices 9 and 10. Therefore, the subset of vertices $\{5, 6, 8, 9, 10\}$ (i.e., the subgraph bounded by the red box) receive graph changes and their label embedding require to be updated. However, updating the position of vertex 6 in the latent space might cause the change of that of vertex 2, or even all of the remaining vertices. It is unclear to what extent we should prorogate those changes: Too aggressive leads to an entire update (equivalent to the recomputing) while too conservative leads to great loss in accuracy.
\end{example}

\vspace{+2mm}
\noindent \textbf{Overview of the incremental algorithm.} We develop an incremental algorithm based on the mean field theory, which assumes that the effect of all the other individuals on any given individual is approximated by a single averaged effect. Hence we choose the candidate set of vertices based on how much they differ from the averaged effect.  The overall process is outlined in Algorithm~\ref{alg:Incremdata}. We first identify a small portion of changes $\Delta V$ as our intial candidate set of vertices $\texttt{cand}$ (Line 3), where $\Delta V$ denote the set of changed vertices (e.g., $\{5, 6, 8, 9, 10\}$ in Fig.~\ref{fig:incre1}), including new and deleted vertices, vertices that have new or deleted edges, and vertices which receive updated ground truth labels. Next, we iteratively perform a conditioned label assignment update for each vertex in the candidate set (Lines 6--7), as well as an update for candidate vertices set $\texttt{cand}$ (Lines 8--10). When updating the candidate vertices, we include one neighbor of an existing candidate vertex into $\texttt{cand}$ only if it satisfies the conditions (Line 9) that are based on the pre-computed value of $w$ and $\delta$ (Lines 4--5). The $w$ exactly denotes the averaged effect of any given individual and $\delta$ denotes the standard deviation of effects of any given individual, and $\theta$ is a confidence level parameter for the trade-off between efficiency and accuracy. We here make an assumption that the propagation matrix $B$ is inherent property and can be accurately estimated by the sampled old data (i.e., $B$ is not changing with new data). The details of candidate vertex update are presented as follows.

\vspace{+2mm}
\noindent \textbf{Update of candidate vertices set.} The set of candidate vertices ($\texttt{cand}$) denotes a subset of vertices, where the label assignment requires update due to graph/label changes. Basically, for each current vertex $u$ in $\texttt{cand}$, we update its label assignment and examine its propagation behavior over its neighbor vertex $v$. Intuitively, the modification of one vertex can cause the relation to its neighbors adjusted, whereas the general behavior between the corresponding two types of vertices should not change much. More specifically, let the $w_{tt'}=\texttt{avg}_{(u,v)\in E_{tt'}}Y(u)^TB_{tt'}Y(v)$ denote the averaged effect of any individual between $t-$ and $t'-$ type, and let the $\sigma_{tt'}=\texttt{std}_{(u,v)\in E_{tt'}}Y(u)^TB_{tt'}Y(v)$ denote the standard deviation of effects between $t-$ and $t'-$ type individuals. If the estimated effect between $u$ and $v$ after adjustment, significantly differs from the averaged effect (i.e., $w_{t(u)t(v)}$) within the same types, we add vertex $v$ into $\texttt{cand}$. The significance is evaluated using a threshold that consists of the confidence parameter $\theta$ and $\sigma_{tt'}$.


\begin{figure}[!hbt]
	\centering
	\includegraphics[width=\columnwidth]{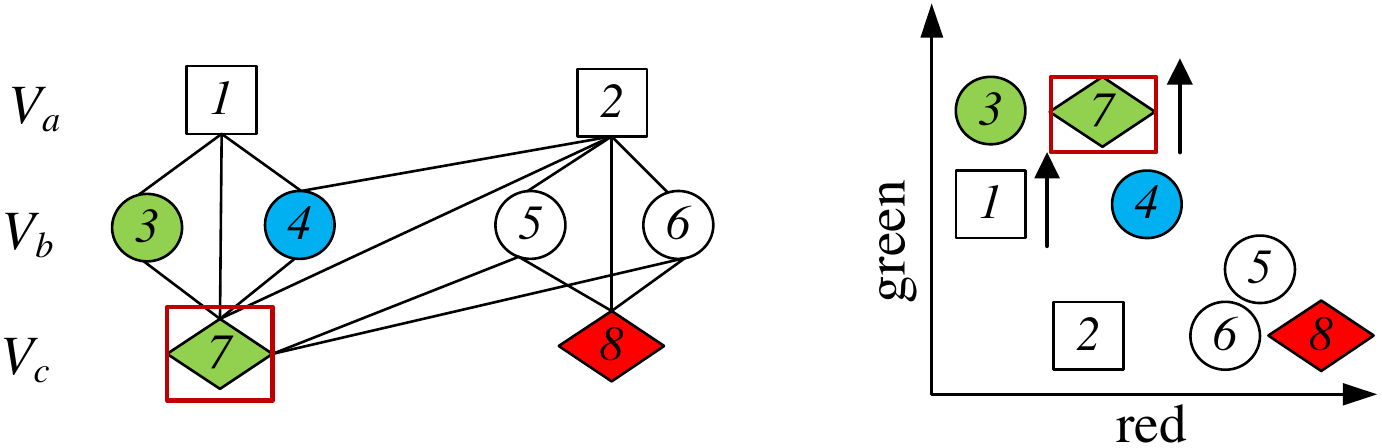}
	\vspace{-5mm}
	\caption{Vertex 7 of the Tripartite graph in Fig.~\ref{fig:tripartiteframework} obtain a new label}\label{fig:incre2}
	\vspace{-0mm}
\end{figure}


\begin{example}
Consider again the example shown in Fig.~\ref{fig:incre1}, the graph changes activate the changes of embedded positions of vertices \{5, 6, 8, 9, 10\} (i.e., vertices in the box). The movement of changed vertices in the label embedding space, consequently causes their neighbor vertices to leave old embedded positions. For example, if vertex 6 is moved further to the red axis, its neighbor vertex 2 might be required to move away from the green axis too. Similarly, in Fig.~\ref{fig:incre2} when vertex 7 receives a new label, vertex 7 are pushed closer to the green axis, which further influences the movement of both vertex 1 and vertex 4. Meanwhile, the new label also strengthens the green label propagation to vertex 1.
\end{example}

\vspace{-1mm}
\noindent\textbf{Confidence level parameter $\theta$ for speed up.} We now discuss the effect of parameter $\theta$ that controls the percentage of neighbors that avoid label update in each iteration (Line 9). A larger value of $\theta$ indicates that more neighbors are filtered out for update, thus leading to a higher confidence that the incremental algorithm is more efficient than a re-computing approach. One nice property of Algorithm~\ref{alg:Incremdata} is that it can bound the number of candidate vertices requiring label assignment update in each iteration. In particular, we present the following theorem that provides the efficiency guarantee for our incremental algorithm.

\vspace{-1mm}
\begin{theorem}\label{theorem:speedup}
In each iteration, the probability $P_c$ that a neighbor of any candidate vertex requires label assignment update is lower bounded by $1-\theta$. That is :
$P_c\leq 1-\theta$.
\end{theorem}
\vspace{-1mm}
\noindent \textbf{Proof (sketch):}  Let $X$ denote a random variable, which represents the value of $Y(u)^TB_{t(u)t(v)}Y(v)$ for any pair of linked vertices $(u, v)$ between $t$-type vertices and $t'$-type vertices. Then $w_{tt'}$ is the average value of $X$, and $\sigma_{tt'}^2$ is the variance of $X$.

Based on Chebyshev's inequality, if $X$ is a random variable with finite expected value $u$ and finite non-zero variance $\sigma^2$. Then for any real number $q > 0$, we have:

\vspace{-2mm}
\begin{equation}
\small{
Pr(|X-u|\geq q\sigma)\leq \frac{1}{q^2}
}
\end{equation}
Let us replace $u$ by $w_{tt'}$, $\sigma$ by $\sigma_{tt'}$, $q$ by $\sqrt{1/(1-\theta)}$, we have:

\vspace{-2mm}
\begin{equation}\label{prob}
\small{
Pr(|X-w_{tt'}|\geq \sqrt{1/(1-\theta)}\sigma_{tt'})\leq (1-\theta)
}
\end{equation}
Eq.~(\ref{prob}) exactly examines the maximum bound of the probability of reaching Line 9 in Algorithm~\ref{alg:Incremdata}. Since the probability of reaching Line 9 in Algorithm~\ref{alg:Incremdata}, is identical to the probability that a neighbor of any candidate vertex requires label assignment update, we complete the proof.\done

In conclusion, the parameter $\theta$ roughly provides a utility that controls the estimated speed up of an incremental algorithm toward a re-computing approach. A larger value of $\theta$ indicates that more neighbors are filtered out for update, thus leading to a higher confidence that the incremental algorithm is more efficient than a re-computing approach. On another hand, a larger value of $\theta$ leads to fewer updates of neighbors and subsequently lower accuracy. From this perspective, the parameter $\theta$ allows us to choose a good trade-off between computational cost and solution accuracy.
%
%

\vspace{-1mm}
\subsection{To Re-compute or Not?}
In the online label inference, we continually receive new data and/or new evidence (labels). Conditioning on the new data and new evidence, we have two
choices: we can recompute the label assignment for all the vertices, using full label inference; or, we can fix some of the previous results, and only update a certain subset of the vertices using an incremental algorithm. To understand the consequences of using an incremental algorithm, we must answer a basic question: how much accuracy loss are incurred and how much speed up are achieved by an incremental algorithm compared to a re-computing approach?

We thus define the utility function for the incremental algorithm $f_I$ as follows:
\begin{equation}\label{equ:utility}
\small{
U(f_I)=u_s\texttt{Gain}(f_I, f_R)-u_a\texttt{Loss}(f_I, f_R)
}
\end{equation}
where $\texttt{Gain}(f_I, f_R)=\frac{|G\cup \Delta G|}{(2-\theta)|\Delta G|}$ is the the computational gain achieved by the incremental algorithm $f_I$ compared to the re-computing algorithm $f_R$, $\Delta G$ is the subgraph induced by the changed vertices $\Delta V$, $\texttt{Loss}(f_I, f_R)$ is the information loss of using incremental algorithm $f_I$ instead of re-computing approach $f_R$, $u_s$ is the reward unit for speed up and $u_a$ is the reward unit for accuracy.

We next examine where the information loss of an incremental algorithm comes. Basically, the accuracy loss of the incremental algorithm comes from two parts: the loss on $V^-$ due to fixing their label assignments, and the loss on $V^+$ due to the approximating label assignments for $V^+$ with fixed previous assigned labels for $V^-$. Therefore, we have:

\vspace{-4mm}
\begin{equation}\label{equ:loss}
\small{
{
\texttt{Loss}(f_I, f_R)=D(Y|f_R, Y|f_I, V^+)+D(Y|f_R, Y_o, V^-)
}}
\vspace{-2mm}
\end{equation}

where $Y|f$ denotes the label assignments using label inference operator $f$, $Y_o$ denotes the previous assigned labels, and $D(Y_1, Y_2, V)$ denotes the label assignment differences of vertices set $V$ between two assignments $Y_1$ and $Y_2$.

Unfortunately, it is non-trivial to examine differences between label assignments by a re-computing and those by an incremental algorithm. This is a chicken and egg paradox: one wants to decide which algorithm to use based on information loss (label assignments differences); while without applying both algorithms, one can not get an accurate understanding about the differences. In order to proceed, we estimate the information loss by examining the similarity between old data and new data, which is inspired by the concept drift modeling~\cite{Gama:2014:SCD:2597757.2523813,vzliobaite2010learning} for supervised learning. The intuition is that if the distribution of new data significantly varies from that of old data, an incremental update results in higher information loss; while the new data are very similar to old data, an incremental update leads to much less information loss. Specifically, we use the graph proximity heuristic defined in Eq.~(\ref{equ:simgraph}) to simulate the similarity between new data and old data, and consequently the information loss, which leads to the following equations:

\vspace{-3mm}
\begin{equation}\label{equ:loss1}
\small{
	\begin{aligned}
	D(Y|f_R, Y|f_I, V^+)=&|\avg_{u\in V^-, v\in V^+}\texttt{sim}(u, v, G\cup \Delta G)\\
	&-\avg_{u\in V^-, v\in V^+}\texttt{sim} (u, v, \Delta G)|
	\end{aligned}
}
\end{equation}
\begin{equation}\label{equ:loss2}
\small{
	\begin{aligned}
	D(Y|f_R, Y_o, V^-)=&|\avg_{u\in V^L, v\in V^-}\texttt{sim}(u, v, G\cup \Delta G)\\
	&-\avg_{u\in V^L, v\in V^-}\texttt{sim} (u, v, G)|
	\end{aligned}
}
\end{equation}

Though information loss might happen when using an incremental algorithm, much less time is consumed in computation compared to any re-computing operation, especially when the percentage of changes is very small.

\section{Experiments}
\subsection{Datasets and Settings}
We evaluate the proposed approaches on four real datasets with three different classification tasks: sentiment classification, topic classification and rating classification. Among the four datasets, Prop 30 and Prop 37 are two tripartite graphs created from 2012 November California Ballot Twitter Data~\cite{ZhuGCL14}, each of which consists of tweet vertices, user vertices, and word vertices. The label classes are sentiments: positive, negative, and neutral. The PubMed dataset~\cite{PubMed} is represented as a tripartite graph with three types of vertices: papers, reviewers, and words. Each paper/reviewer is associated with multiple labels, which denote the set of subtopics. The MovieLen dataset~\cite{MovieLenData} represents the folksonomy information among users, movies, and tags. Each movie vertex is associated with a single class label. Here we use three coarse-grain rating classes, ``good", ``neutral", and ``bad" as the ground truth labels. A short description of each dataset is summarized in Table~\ref{tab:data}.

Let MRG/ARG denote the multiplicative/additive rule update with graph heuristic initialization. We compare our approaches with three baselines: GRF~\cite{Zhu03semi-supervisedlearning}, MHV~\cite{Ding:2009:LPK:1726586.1727013} and BHP~\cite{GatterbauerVLDB2015}. GRF is the most representative traditional label propagation algorithm (i.e., no $B$ matrices or $B$ matrices are identity matrices), MHV is selected as a representative method that supports vertex-level heterogeneity ($B$ matrices are diagonal), and BHP denotes the label propagation algorithm that allows propagation-level heterophily and utilizes a single matrix $B$. For all of these approaches, we begin with the same initialized state, and we use the same regularizations/or no regularizations for all approaches.  Note that our goal in this paper is not to justify the performance of semi-supervised learning for different classification tasks (various surveys have justified the advantage of semi-supervised learning with fewer labeled data), but rather to propose a better semi-supervised label propagation algorithm for tripartite graphs. Therefore, we do not compare our approaches with other supervised methods such as support vector machine.

We evaluate the effectiveness of each approach in terms of classification accuracy. Specifically, we select [1\%, 5\%, 10\%] of vertices with ground truth labels as the set of seed labeled vertices, and then run different label propagation algorithms over the entire network to label the remaining vertices. Note that the selection of seed labeled vertices is not the focus of this work, and thus we simply adopt the degree centrality to select the top [1\%, 5\%, 10\%] of vertices as seed nodes. For both single- and multi-class labels, we assign label $l_i$ to a vertex $u$ if $Y(u,l_i)> 1/k$ and validate the results with ground truth labels. We represent classification results as a contingency matrix $A$, with $A_{ij}$ for $i, j\in L=\{l_1,\cdots, l_k\}$ where $k$ is the number of class labels, and $A_{ij}$ is the number of times that a vertex of true label $l_i$ is classified as label $l_j$. With the contingency matrix $A$, the classification accuracy is defined as: $\texttt{Accuracy}=\frac{\sum_{i}A_{ii}}{\sum_{ij}A_{ij}}$.

The classification accuracy can not well evaluate the performance if label distribution is skewed. Therefore, we also use Balanced Error Rate~\cite{read2007automatic} to evaluate the classification quality. The balanced error rate is defined as $\texttt{BER}=1-\frac{1}{k}\sum_{i}\frac{A_{ii}}{\sum_{j}A_{ij}}$.

\begin{table}[!ht]
	\caption{{The statistics of datasets. ``single label" denotes whether each vertex can be associate with one or more labels; ``max label" denotes the class label $l_m$ that has maximum number of vertices and ``\% max label" denotes the percentage of vertices that have ground truth label $l_m$.} }\label{tab:data}
		\small{
			\begin{tabular}{|c|c|c|c|c|}
				\hline
				Data&Prop 30&Prop 37&MovieLen&PubMed\\
				\hline
				\# nodes&23,025&62,383&26,850&36,813\\
				\hline
				\# edges&168,131&542,028&168,371&263,085\\
				\hline
				\# classes&2&2&3&19\\
				\hline
				single label&Yes&Yes&Yes&No\\
				\hline
				\% $n_a$&3.6\%&3.1\%&14.9\%&0.3\%\\
				\% $n_b$&45.7\%&54.5\%&56.8\%&0.6\%\\
				\% $n_c$&50.7\%&42.4\%&28.3\%&99.1\%\\
				\hline
				\% max label&67\%&90\%&67\%&23\%\\
				\hline
			\end{tabular}}
		\end{table}

All the algorithms were implemented in Java 8 in a PC with i7-3.0HZ CPU and 8G memory.

\subsection{Static Approach Evaluation}
In this section, we evaluate the performance of Algorithm~\ref{alg:framework} in terms of convergence, efficiency and classification accuracy. 

\begin{figure}[!t]
  \centering
  \includegraphics[width=0.85\columnwidth]{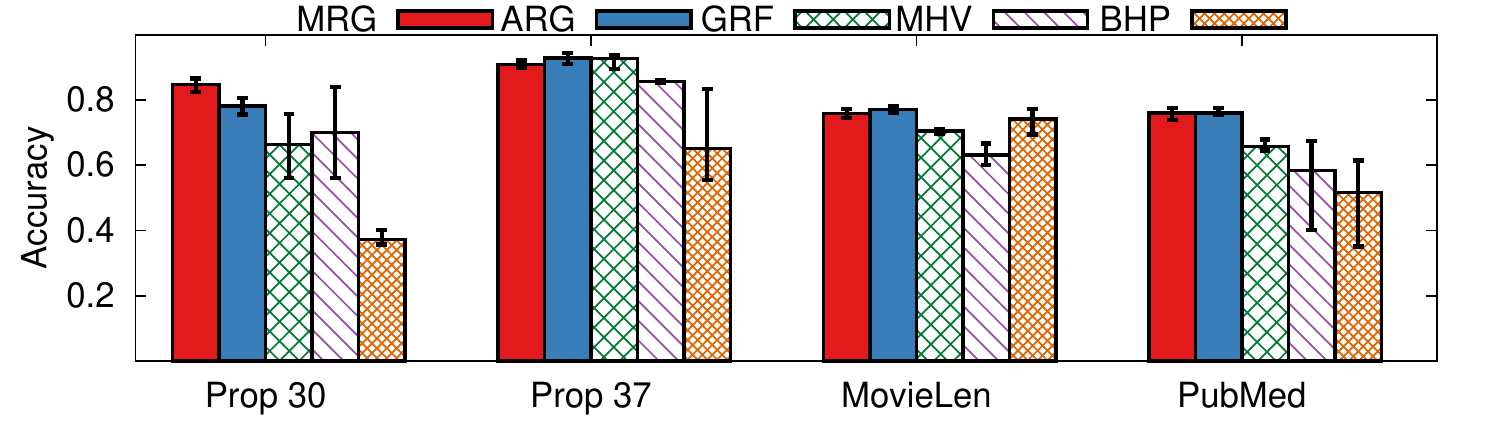}
  \includegraphics[width=0.85\columnwidth]{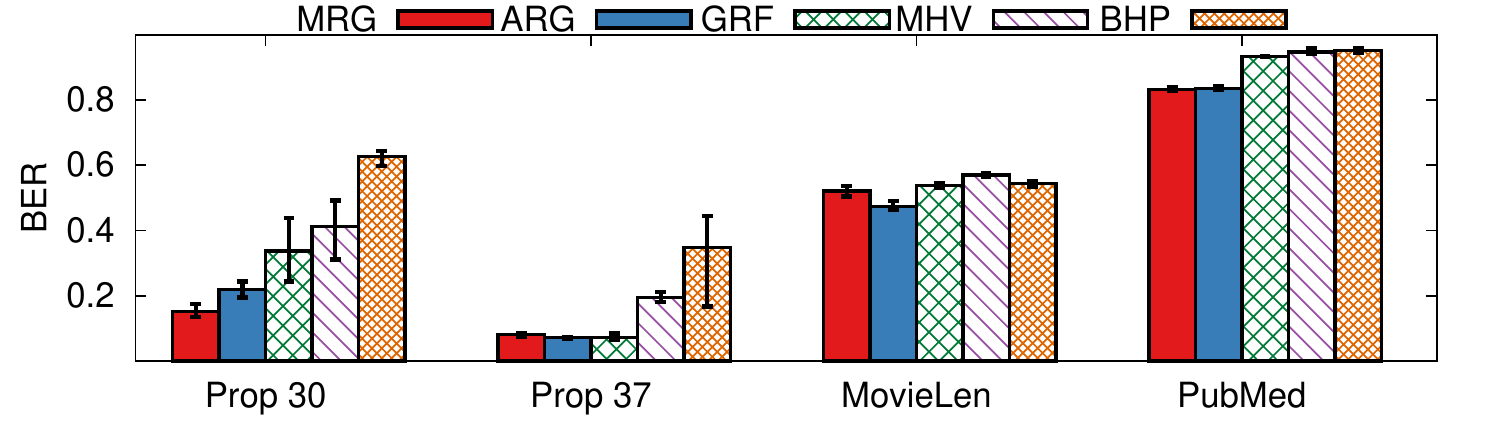}
  \caption{Classification quality comparisons. The higher accuracy (the lower balanced error rate), the better quality.}\label{fig:accuracy}
\end{figure}

\begin{ques}
\textbf{Accuracy}: How does Algorithm~\ref{alg:framework} perform compared to the baselines?
\end{ques}
\begin{myans}
\small{
\emph{
Algorithm~\ref{alg:framework} outperforms the baselines in terms of classification accuracy and balanced error rate when data exhibit heterophily and/or with multi-class labels.}}
\end{myans}

\begin{figure}[!t]
  \centering
  \subfigure[Prop 30, MRG]{\includegraphics[width=0.15\columnwidth]{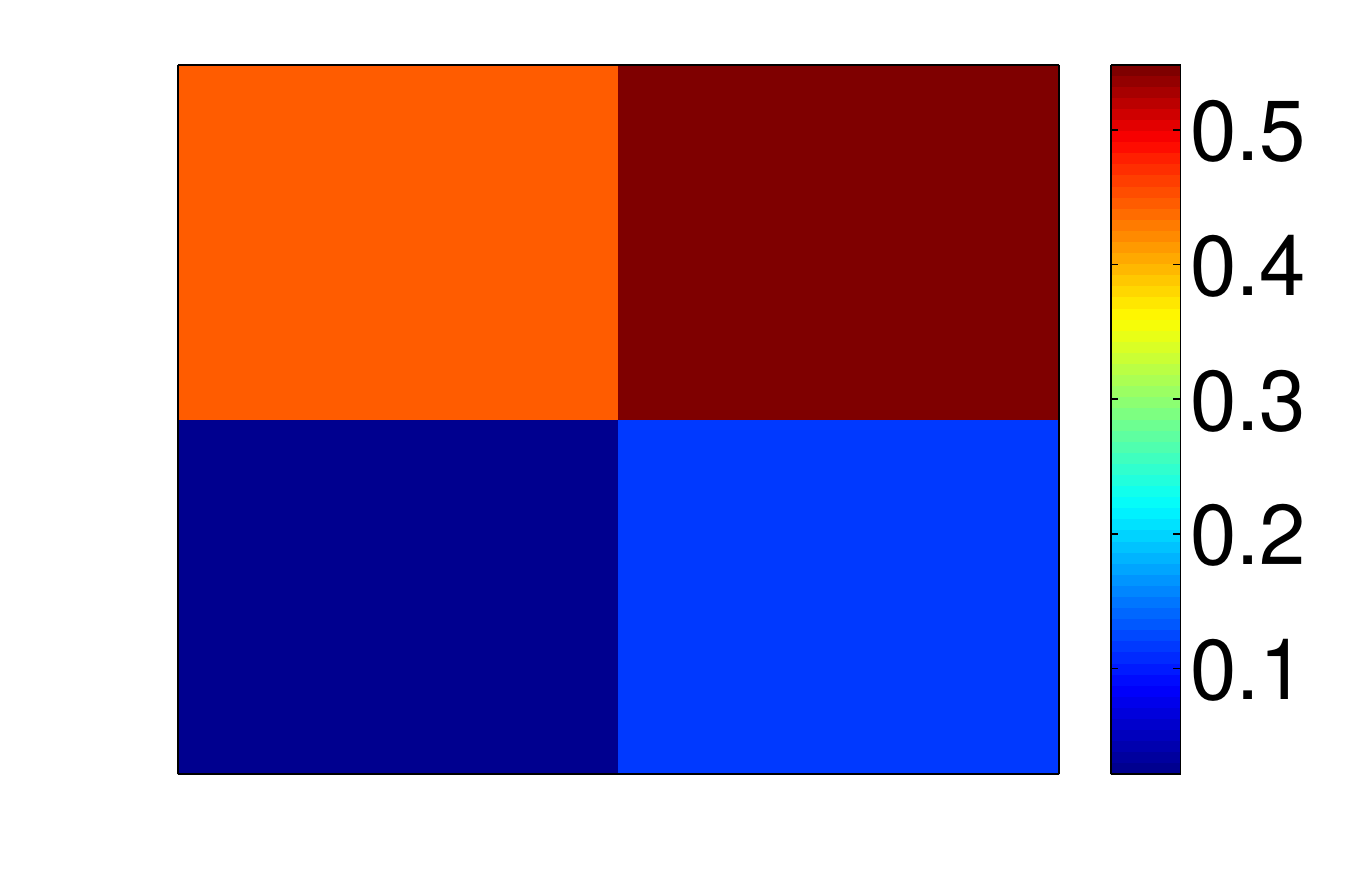}
  \includegraphics[width=0.15\columnwidth]{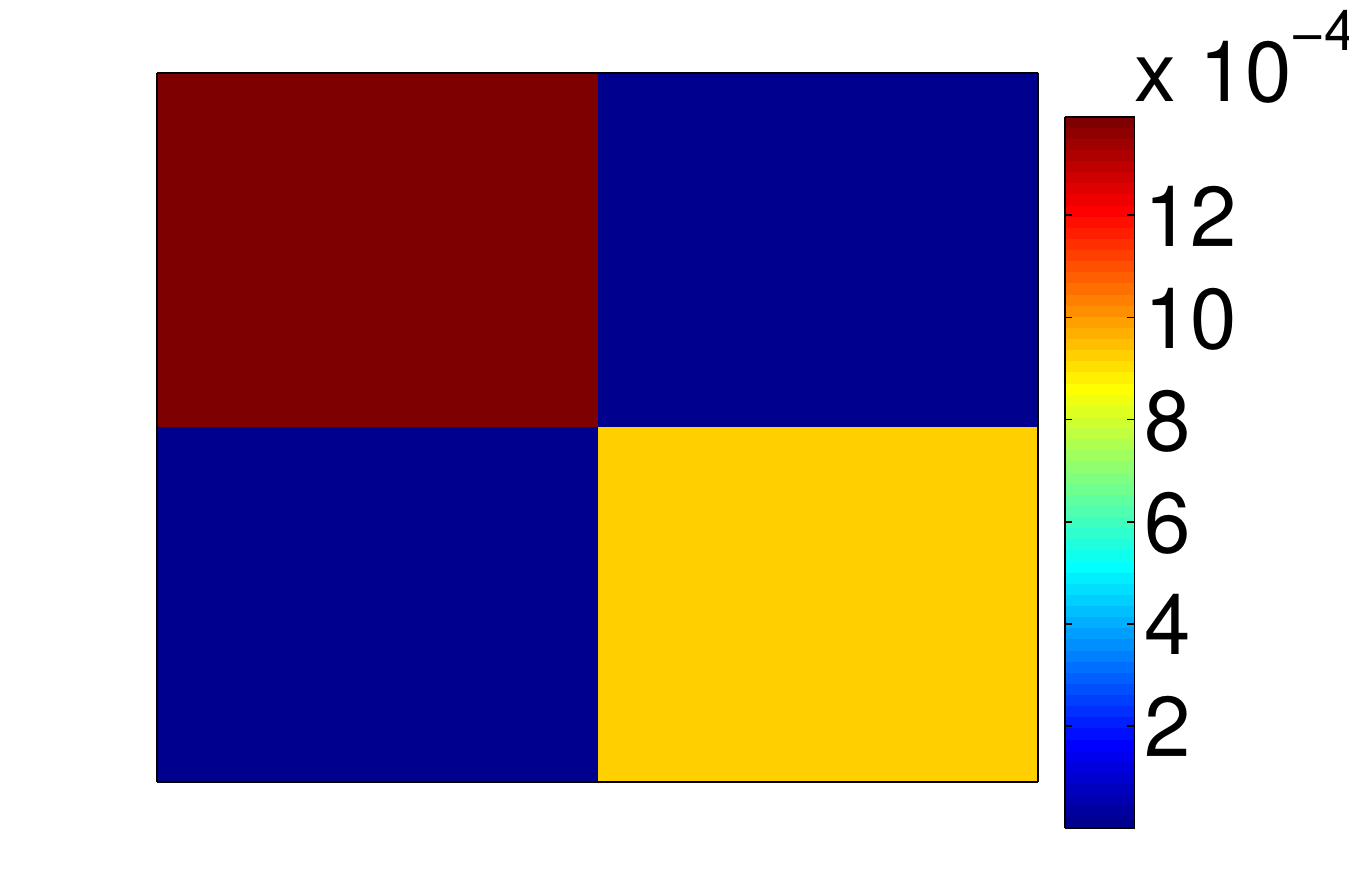}
  \includegraphics[width=0.15\columnwidth]{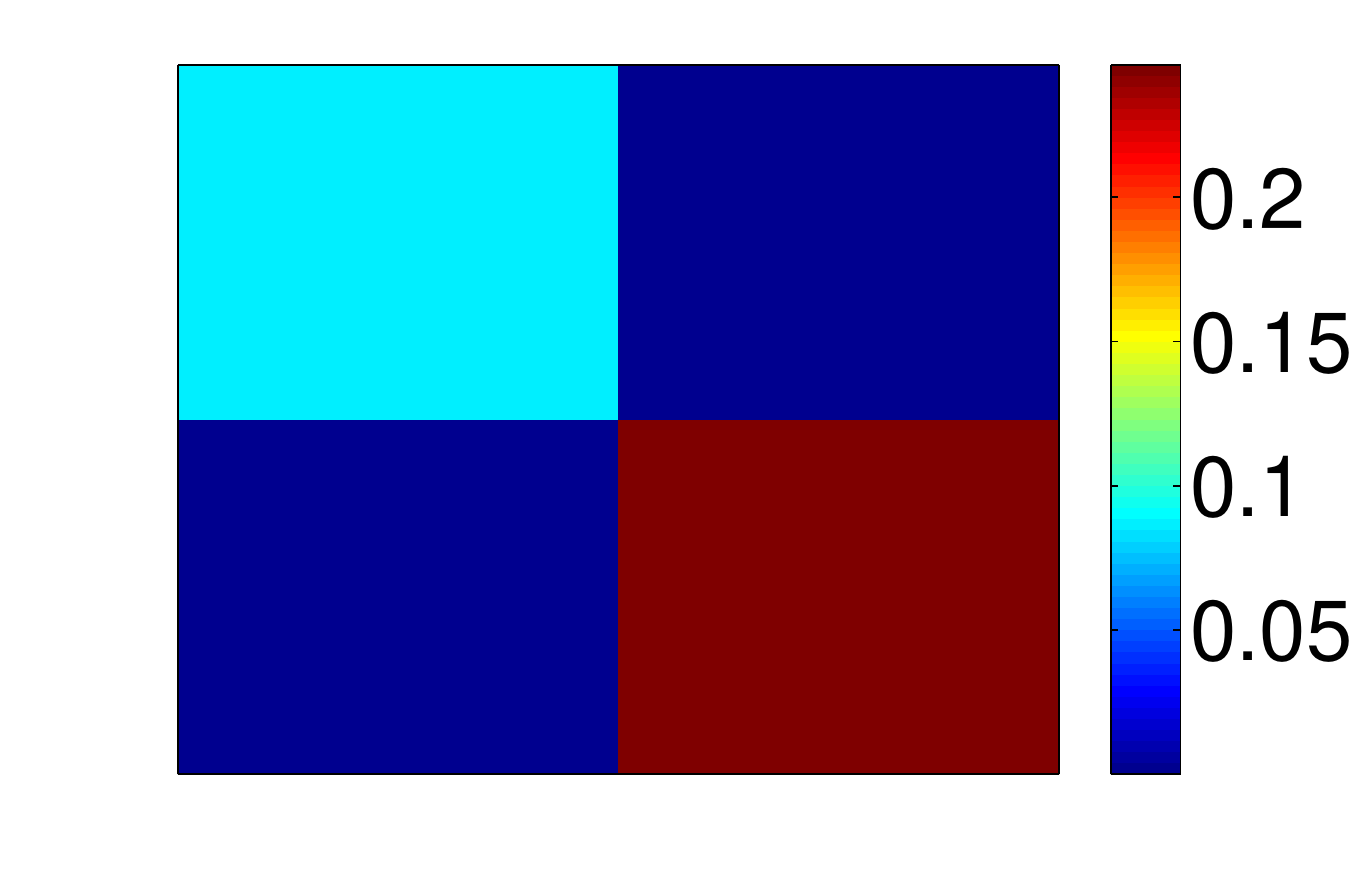}}
  \subfigure[Prop 37, ARG]{\includegraphics[width=0.15\columnwidth]{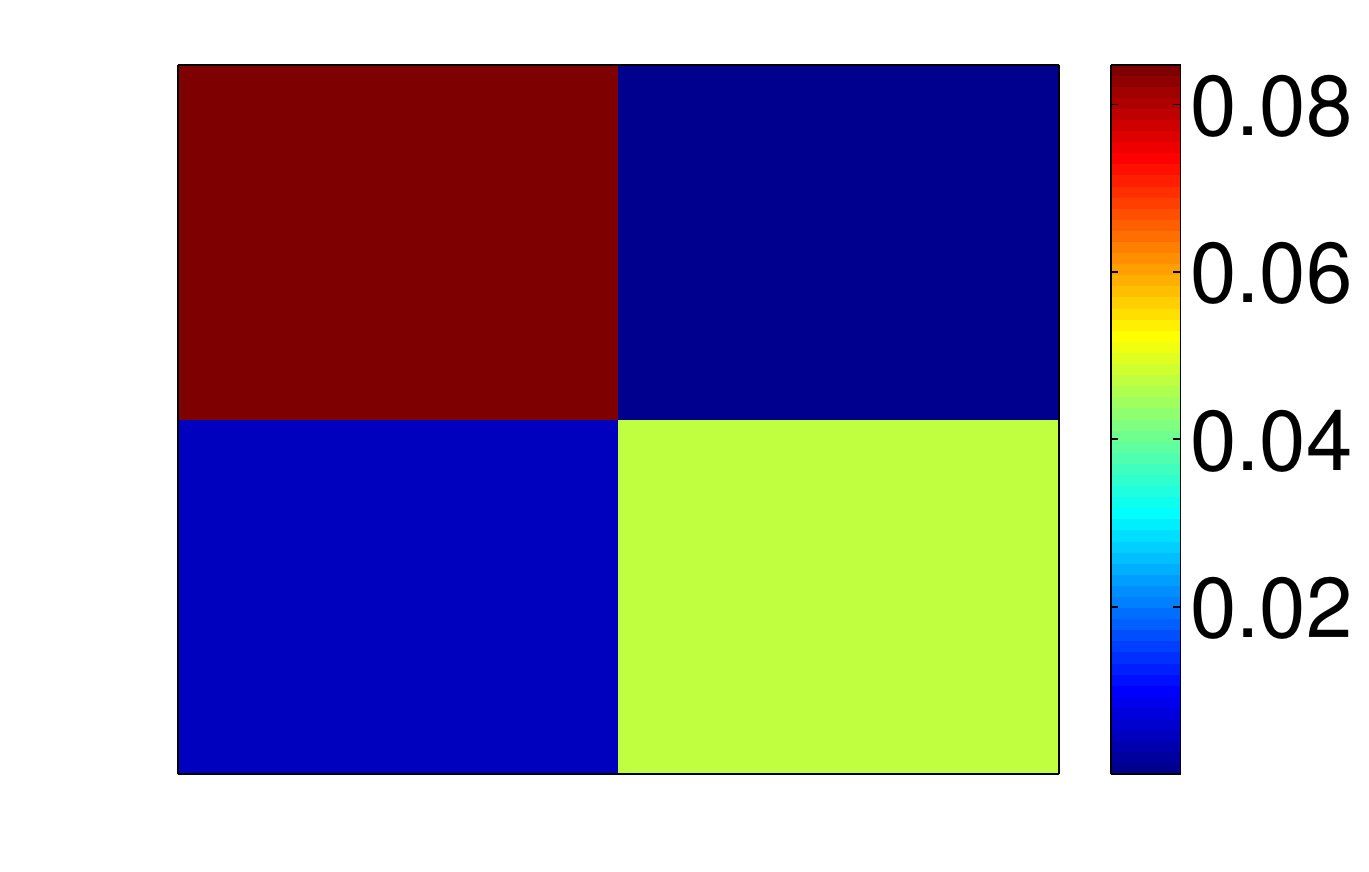}
  \includegraphics[width=0.15\columnwidth]{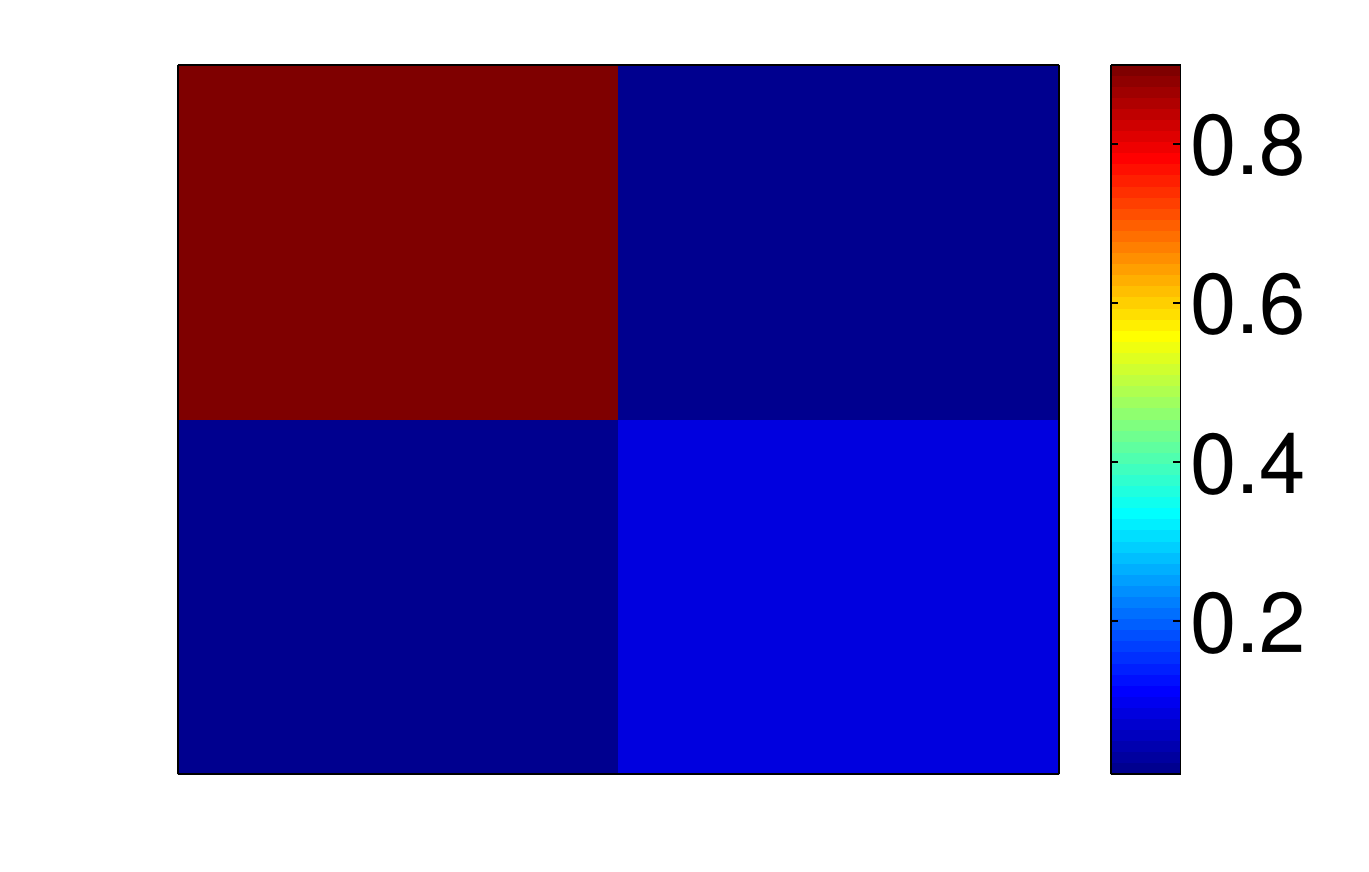}
  \includegraphics[width=0.15\columnwidth]{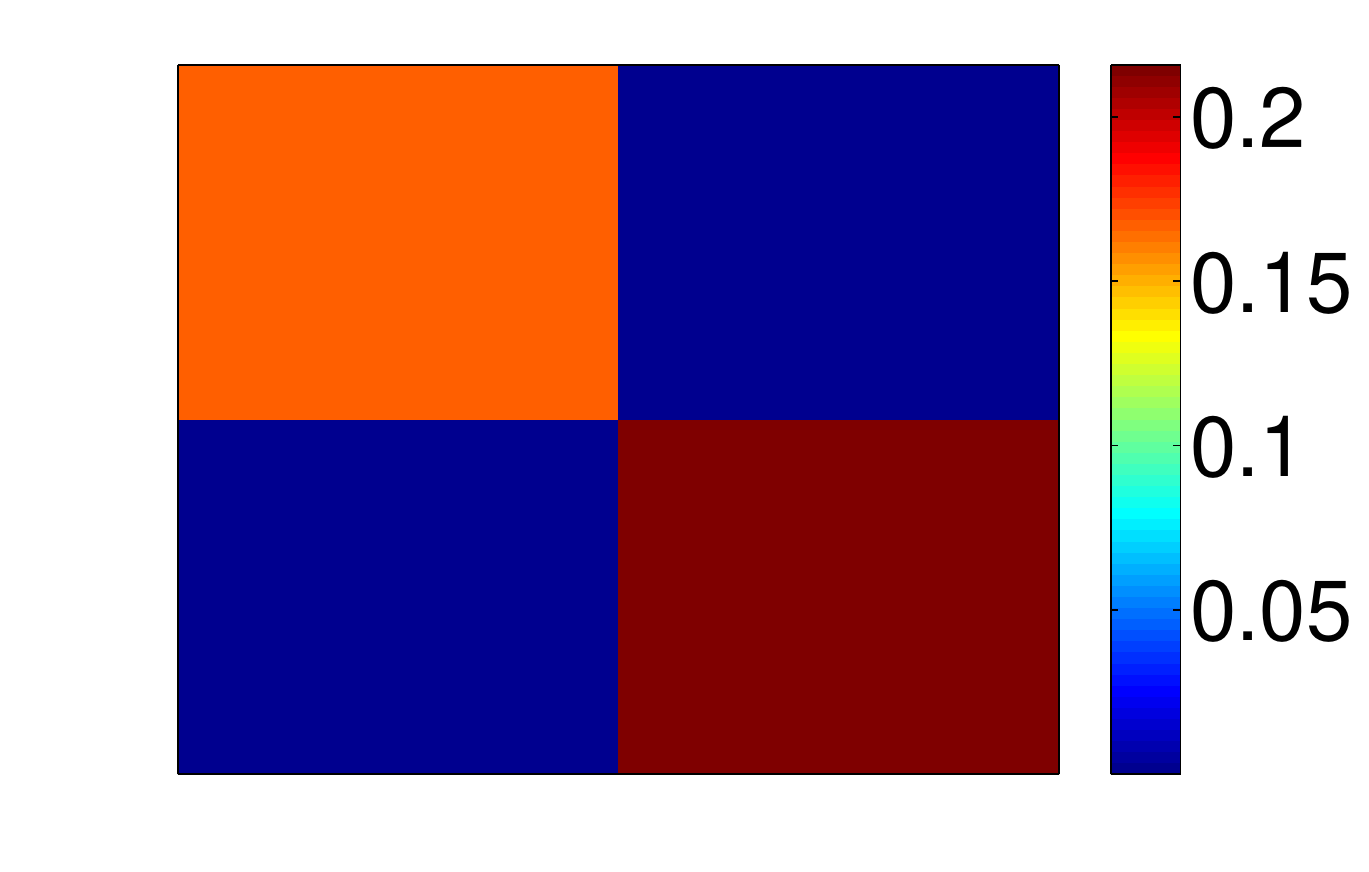}}
    \subfigure[MovieLen, ARG]{\includegraphics[width=0.16\columnwidth]{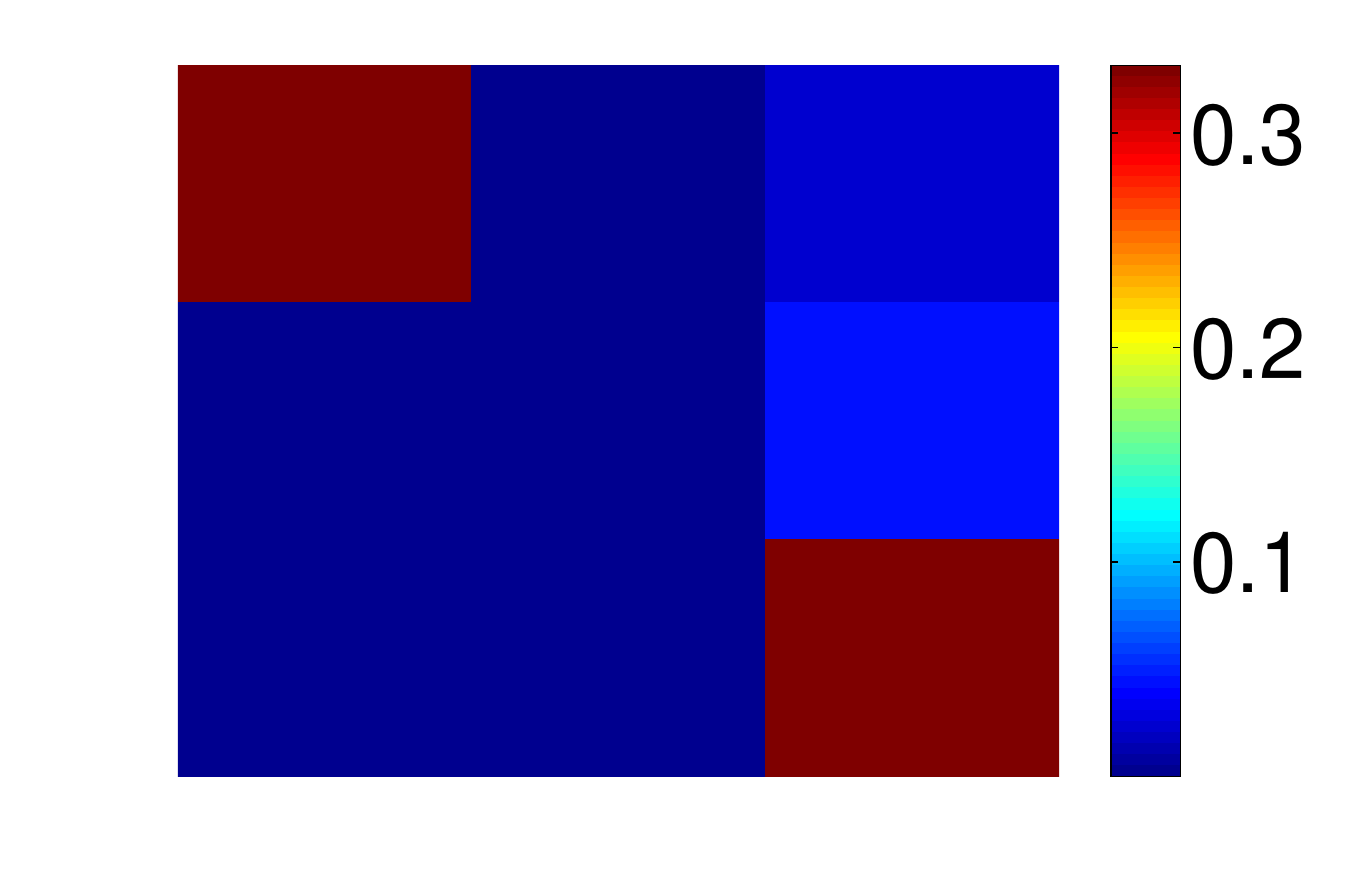}
  \includegraphics[width=0.153\columnwidth]{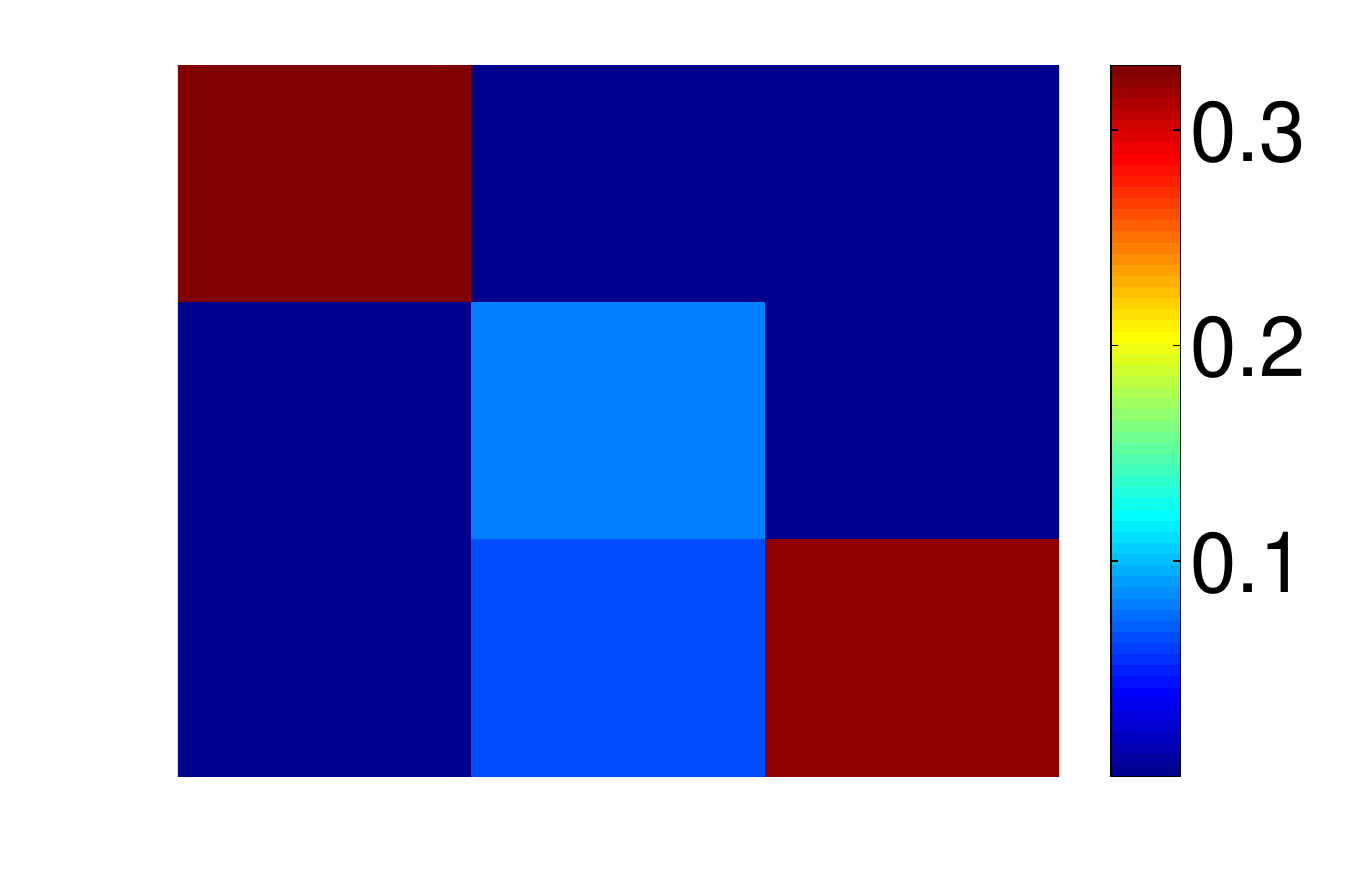}
  \includegraphics[width=0.153\columnwidth]{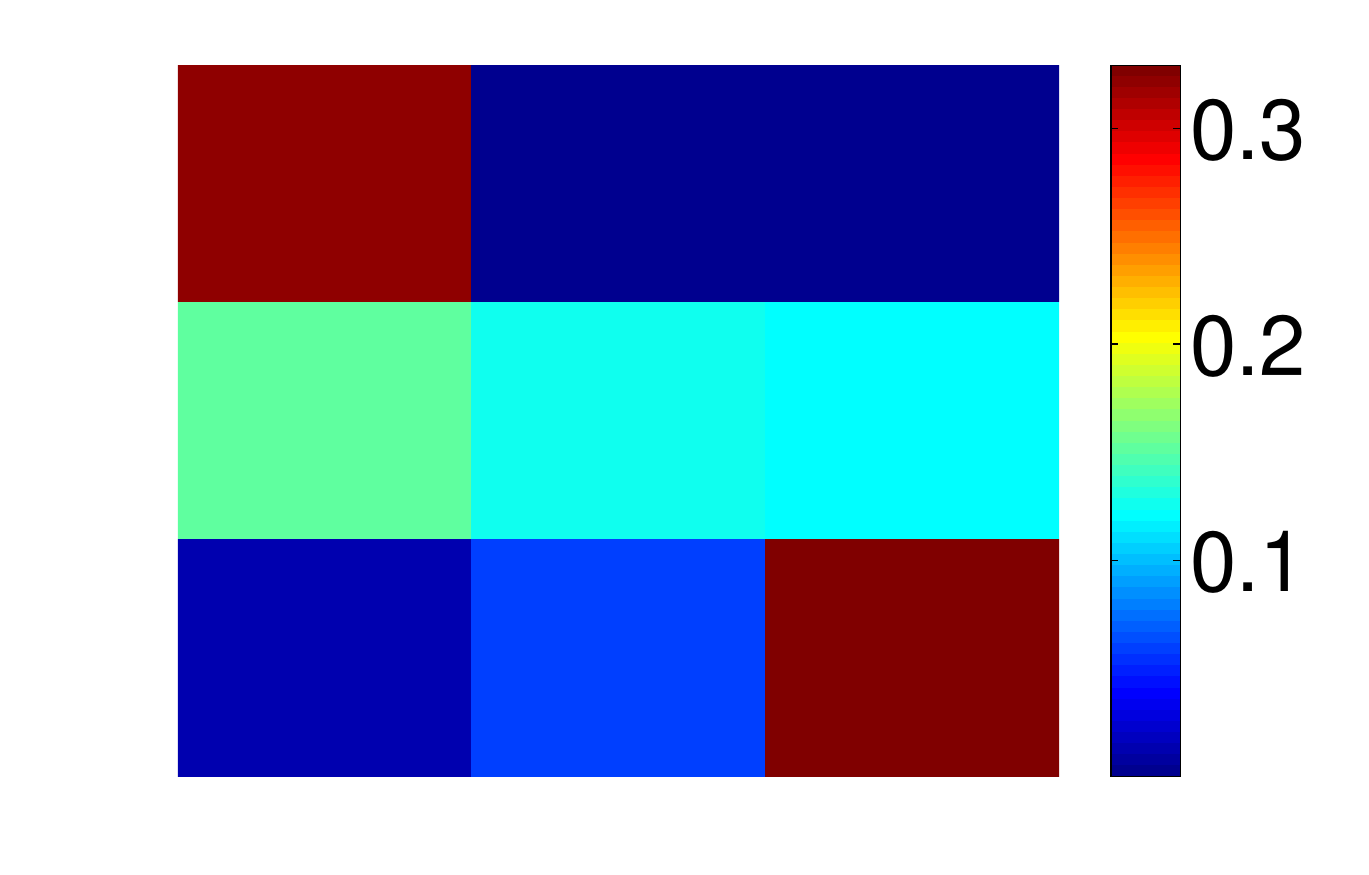}}
  \subfigure[PubMed, MRG]{\includegraphics[width=0.153\columnwidth]{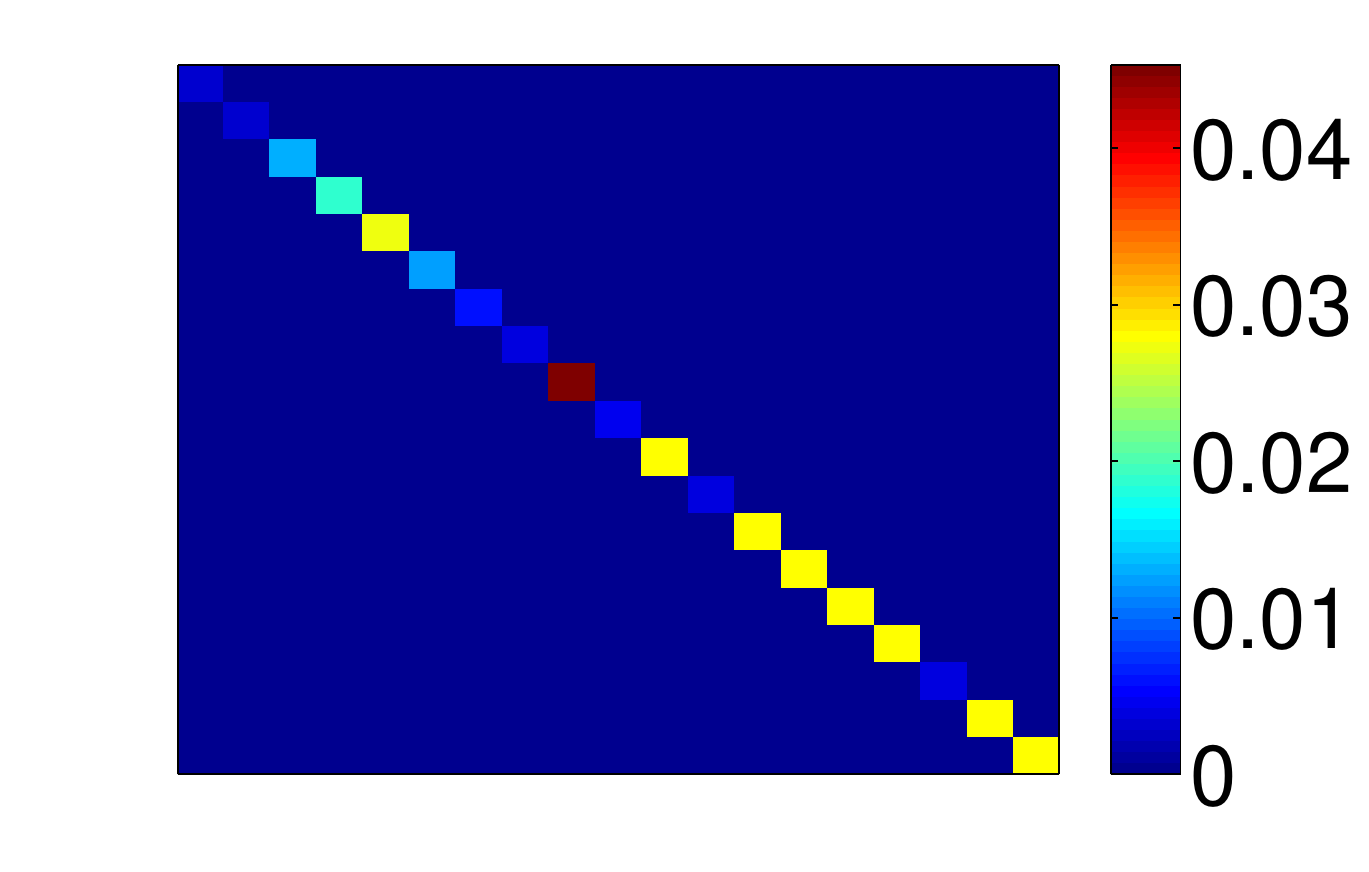}
  \includegraphics[width=0.153\columnwidth]{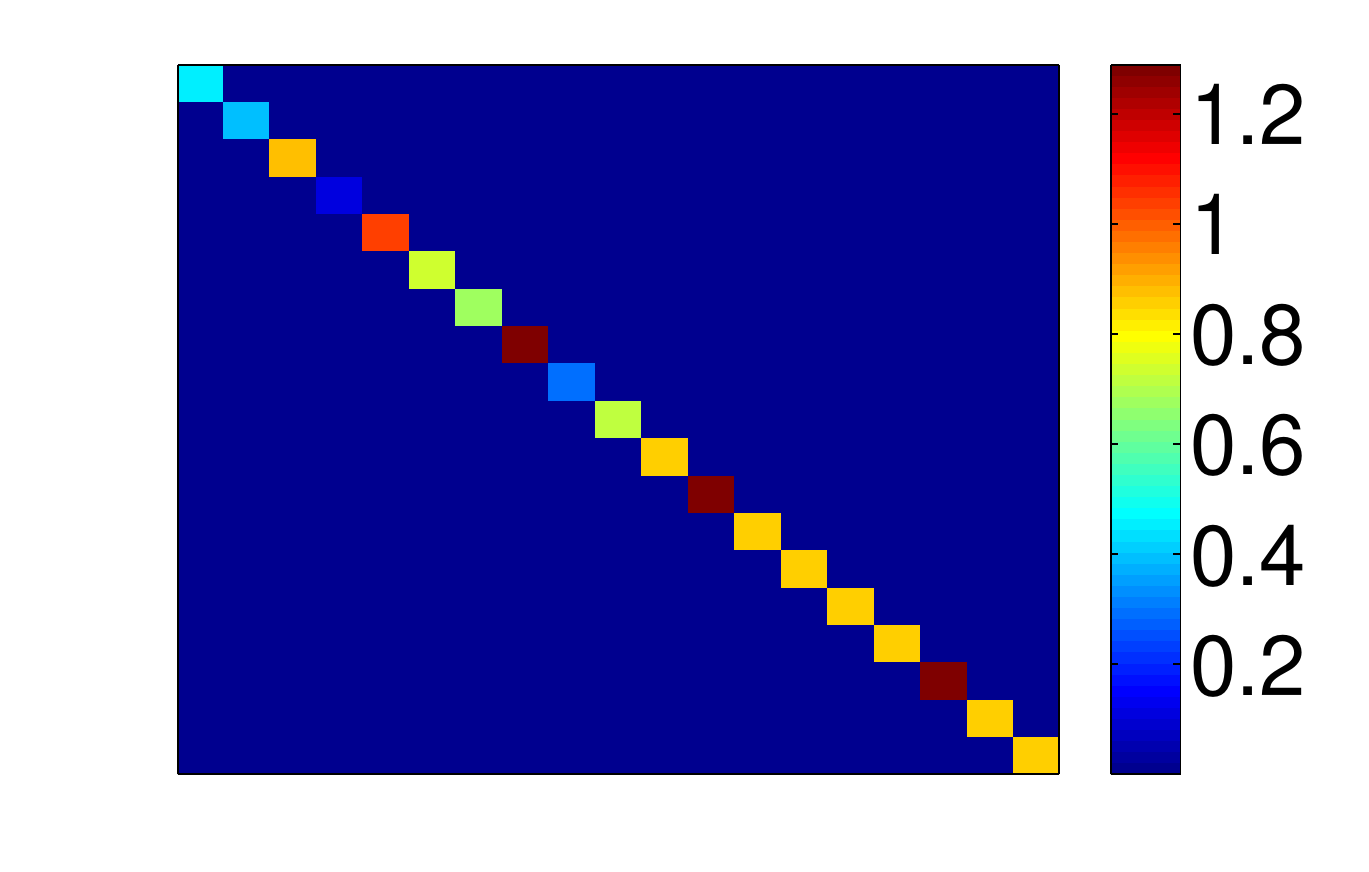}
  \includegraphics[width=0.153\columnwidth]{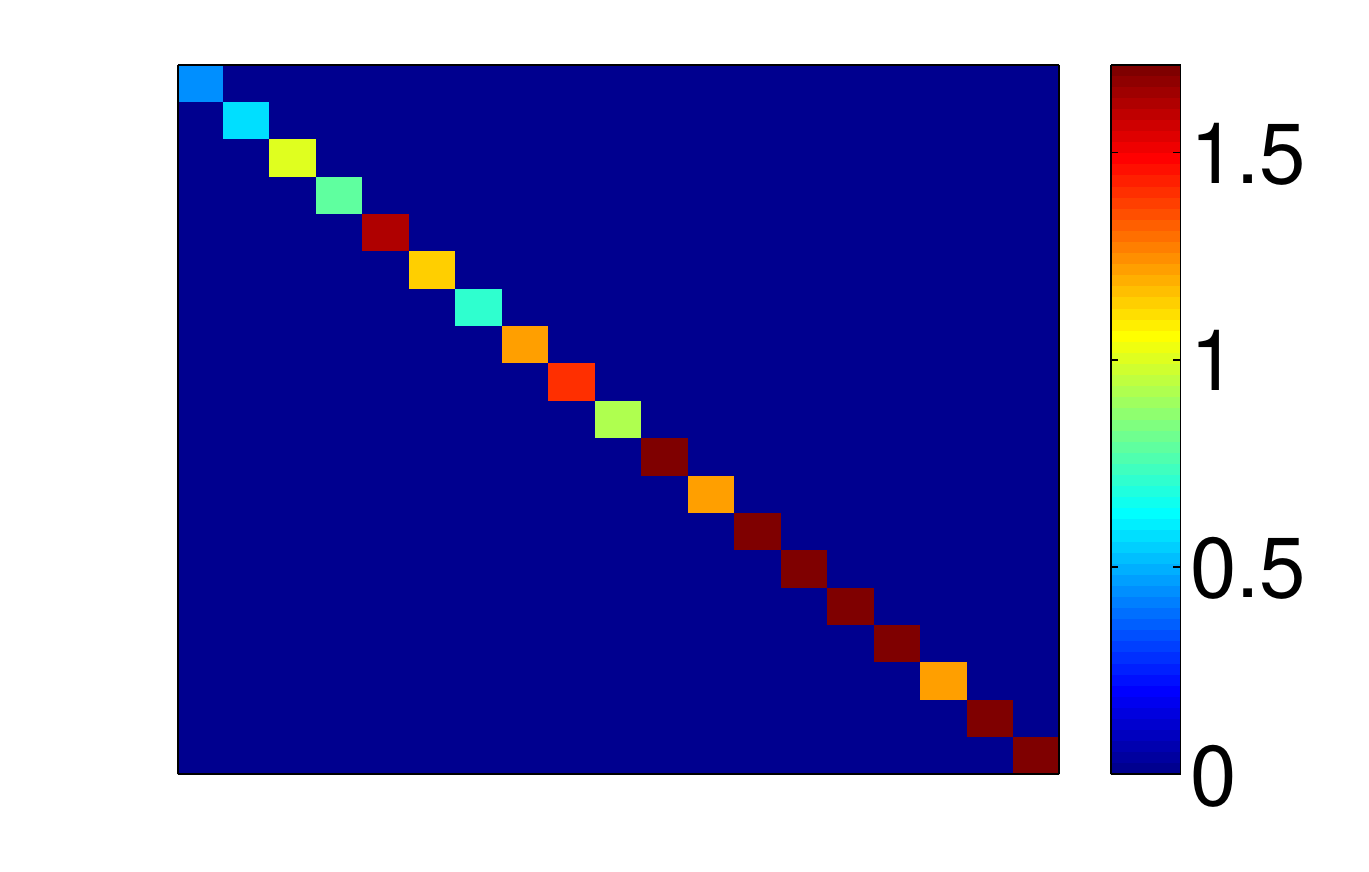}}
  \caption{The automatically learned propagation matrices $B$ by the better algorithm between ARG and MRG (Best viewed in color).}\label{fig:B}
\end{figure}

\vspace{0.1cm}
Fig.~\ref{fig:accuracy} reports the classification accuracy and balanced error rate comparisons for all the approaches. Here we set the convergence tolerance value to $10^{-6}$. The parameter $\beta$ is set to 5. This is because in our preliminary experiments, we notice that the accuracy increases when $\beta$ is increased from 1 to 5, but after that, increasing $\beta$ does not lead to significant changes in accuracy. Clearly, our approaches MRG and ARG, performs much better than the other approaches on Prop 30, MovieLen, PubMed, and perform similarly with GRF on Prop 37. The results validate the advantage of inferred $B$ matrix in terms of supporting vertex-level heterogeneity, propagation-level heterogeneity and the multi-class label propagation.

For example, Prop 37 is about labeling genetically engineered foods, and the majority of people have positive attitude. As shown in Fig.~\ref{fig:B} (b), all of the $B$ matrices learned by ARG are diagonal matrices, which illustrates that the link formation exactly follows homophily assumption. Hence, on Prop 37, forcing $B$ matrices as identity matrices such as GRF obtains comparable quality with our approaches. Moreover, although both PubMed (see Fig.~\ref{fig:B} (d)) and Prop 37 exactly follow homophily assumption, the proposed approaches perform better than other approaches on PubMed. This is because PubMed has multi-class labels and our approaches well support multi-class label propagation compared to other approaches. On Prop 30 and MovieLen, the $B$ matrices are a mixture of different forms of matrices. Under this situation, our approaches MRG and ARG perform better than all the approaches including BHP (using a single arbitrary B matrix). 



\begin{figure}[!t]
  \centering
  \includegraphics[width=0.8\columnwidth]{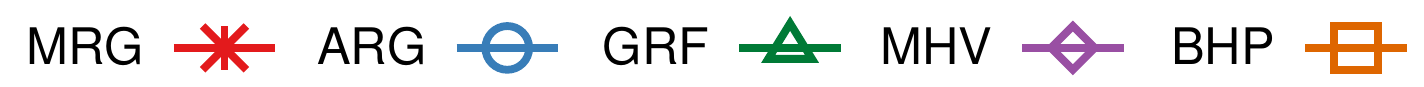}
  \subfigure[Prop 30]{\includegraphics[width=0.49\columnwidth]{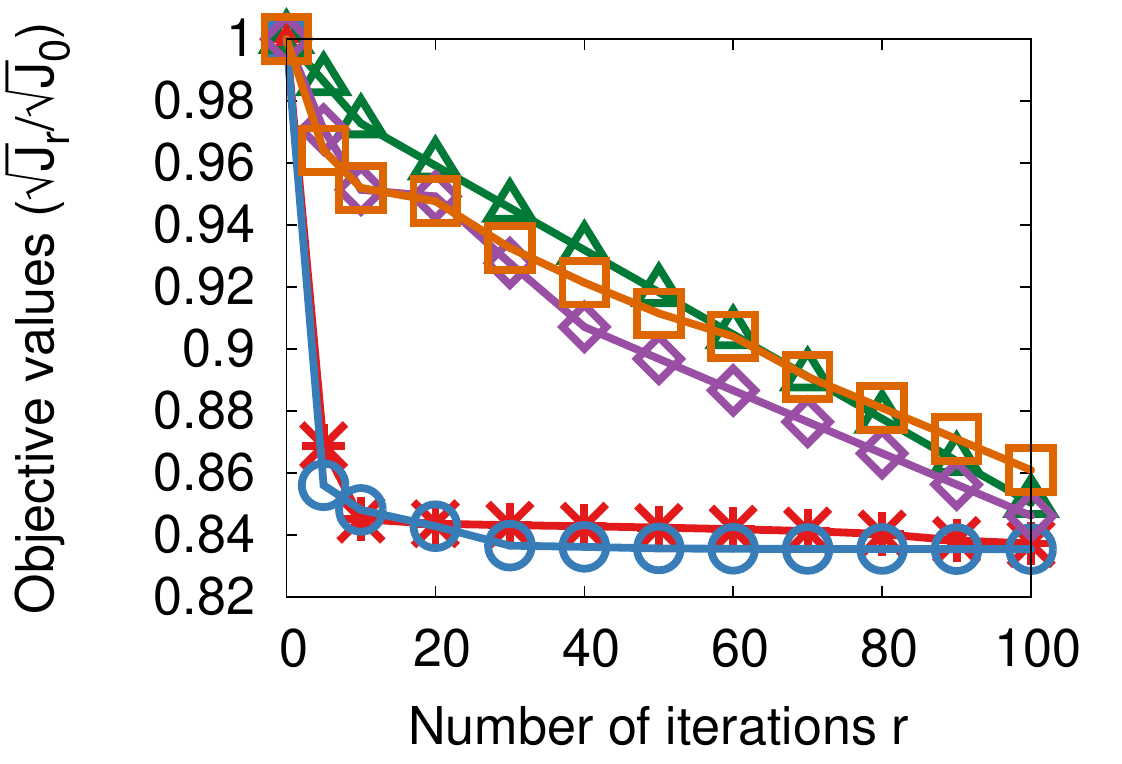}}
  \subfigure[Prop 37]{\includegraphics[width=0.49\columnwidth]{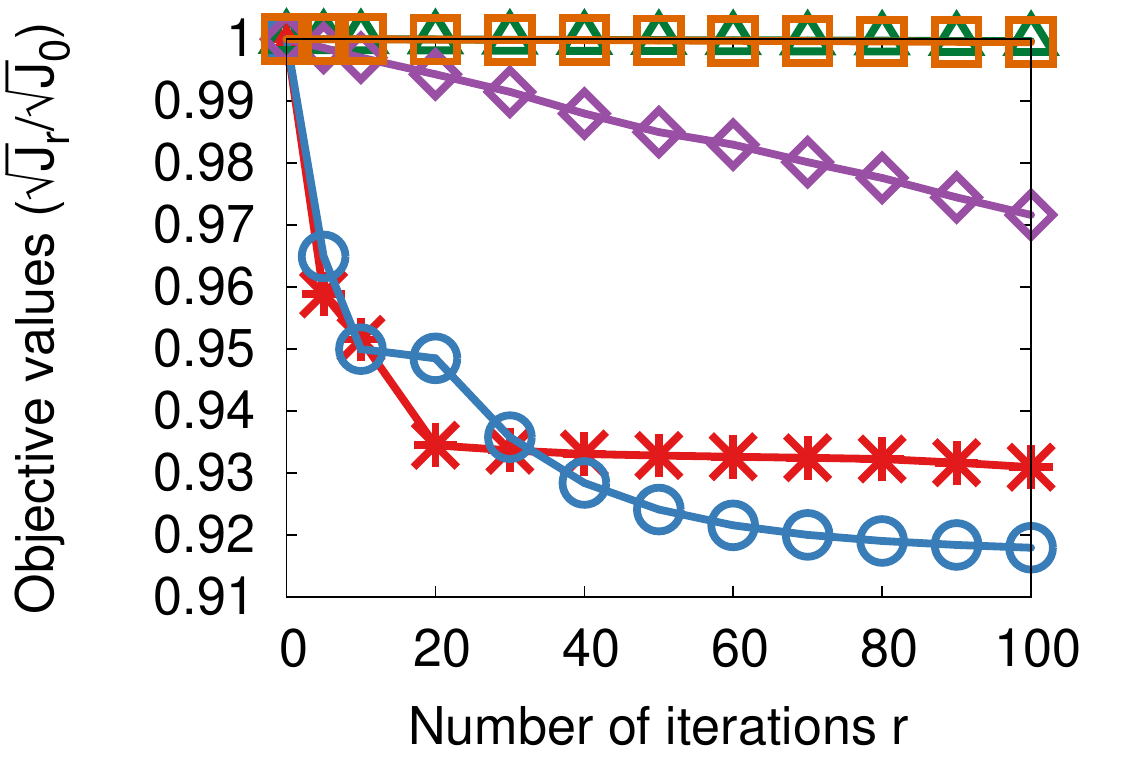}}
  \subfigure[MovieLen]{\includegraphics[width=0.49\columnwidth]{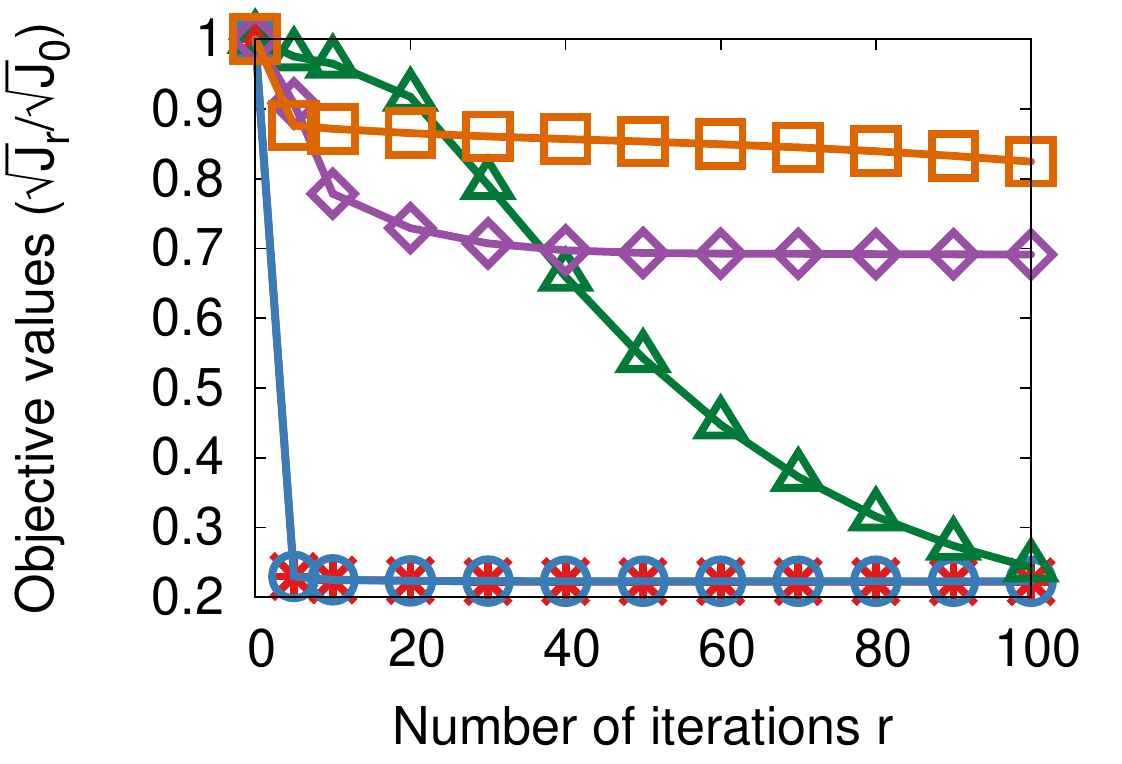}}
  \subfigure[PubMed]{\includegraphics[width=0.49\columnwidth]{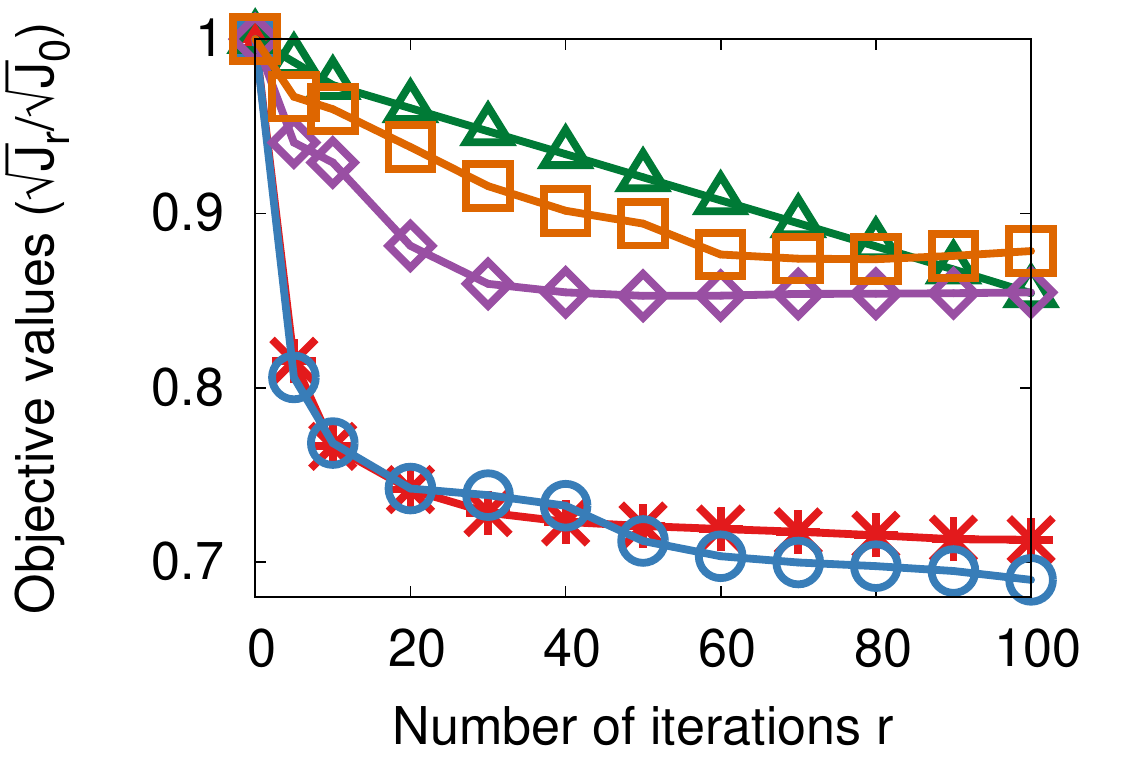}}
  \caption{Convergence comparison. $y$ axis: the ratio between $\sqrt{J_r}$ and $\sqrt{J_0}$, where $J_r$ is the objective value for $r^{th}$ iteration and $J_0$ is the objective value of initialization.}\label{fig:converge}
\end{figure}

\begin{ques}
\textbf{Convergence}: Do multiplicative and additive rules in Algorithm~\ref{alg:framework} converge?
\end{ques}
\begin{myans}
{
\emph{
Both multiplicative and additive rules are guaranteed to converge into local optima and their convergence rate in terms of objective values are much faster than the baselines.}
}
\end{myans}

\vspace{0.1cm}
Instead of fixing convergence tolerance value, now we fix the maximum iteration number to 100, and validate the convergence performance of the proposed algorithms. Fig.~\ref{fig:converge} shows that the objective values are non-increasing using both multiplicative rules and additive rules. In addition, the proposed algorithms decrease the objective value much faster than other algorithms.

\begin{figure}
  \centering
  \includegraphics[width=0.9\columnwidth]{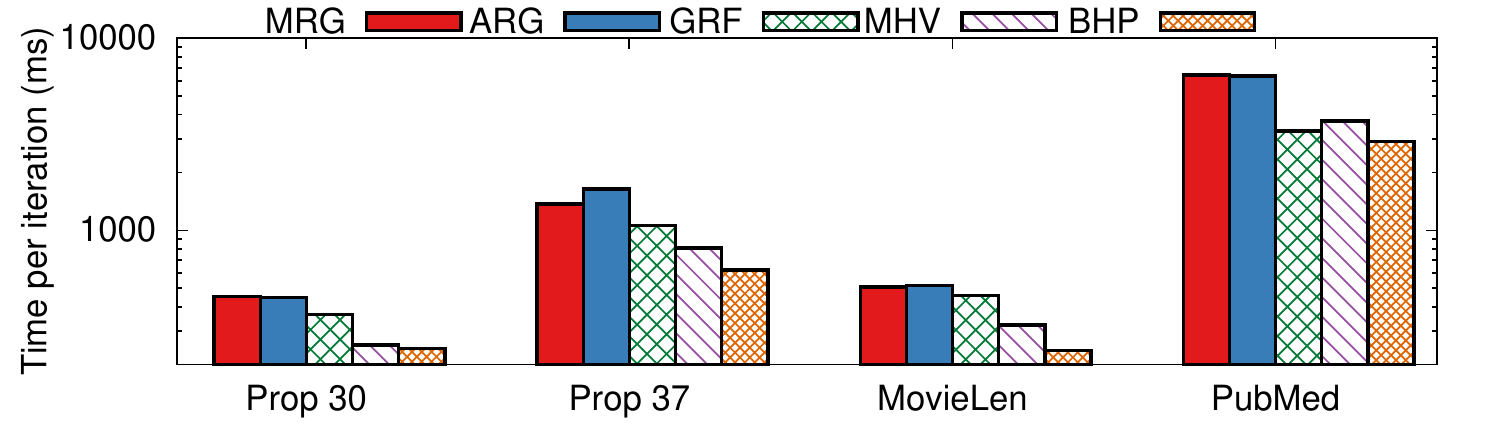}
  \includegraphics[width=0.9\columnwidth]{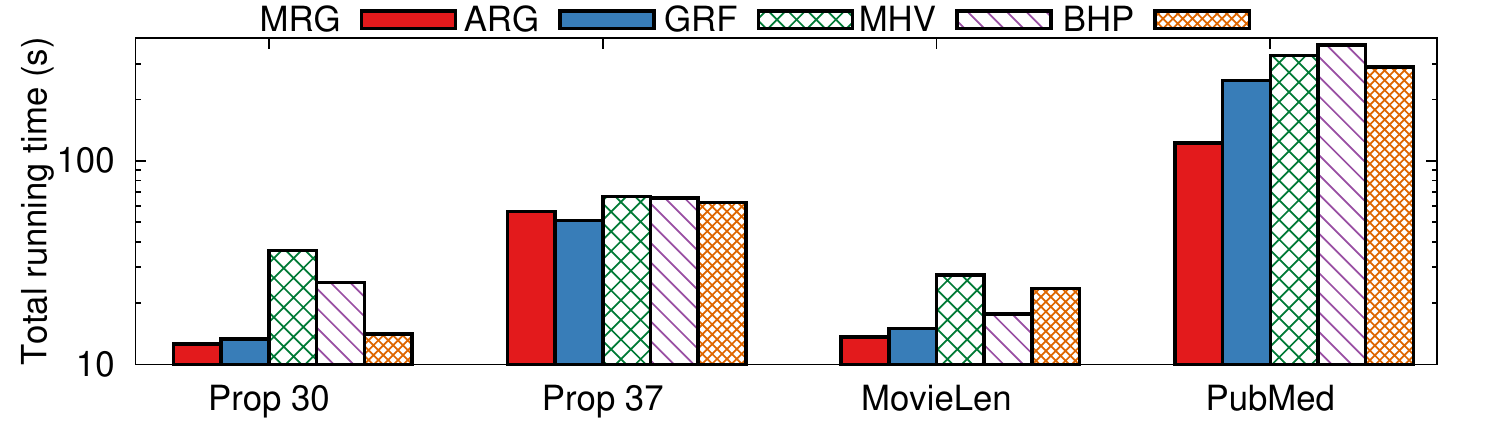}
  \caption{Efficiency comparison. Note that time per iteration is in millisecond scale and total running time is in second scale.}\label{fig:time}
\end{figure}

\begin{ques}
\textbf{Efficiency}: Are the proposed algorithms scalable?
\end{ques}
\begin{myans}
{
\emph{
In each iteration, the proposed algorithms MRG and ARG require more computational cost than the baselines. However, since they converge much faster than other approaches, the total running time of the proposed algorithms, are still faster than the baselines.}
}
\end{myans}

\vspace{0.1cm}
We again fix the convergence tolerance value to $10^{-6}$, and report the average running time per iteration and the total running time in Fig.~\ref{fig:time}. Because on average they converge 2-3 times faster than all the baselines, computationally expensive per iteration due to the incurred additional cost for computing the propagation matrices $B$, the proposed algorithms are still very efficient for large-scale data.
\begin{table}[!h]
  \centering
\caption{The effect of graph regularization.}\label{tab:graphreg}
  \begin{tabular}{|c|c|c|c|c|}
    \hline
     & \multicolumn{2}{|c|} {Prop30} &\multicolumn{2}{|c|}{ Prop37} \\
     \hline
     &w&w/o&w&w/o\\
     \hline
 MRG&0.869&0.848&0.935&0.908\\
 \hline
ARG&0.867&0.781&0.942&0.928\\
    \hline
  \end{tabular}
\end{table}
\begin{ques}
\textbf{Regularization}: What is the effect of the graph regularization?
\end{ques}
\begin{myans}
{
\emph{
We validate that graph regularization term is helpful for sentiment classification tasks on Prop 30 and Prop 37.}}
\end{myans}

\vspace{0.1cm}
Intuitively user-to-user graph (e.g., friendship graph) or document graph (e.g., citation graph) will be very helpful for labeling users or documents. Unfortunately, we do not have such graphs for MovieLen and PubMed. Therefore, although our framework is very general and supports various regularization, we only compare the classification accuracy w/o graph regularization on Prop 30 and prop 37 (i.e., the two data sets that have additional user to user re-tweeting graphs). With the additional regularization, the classification accuracy increases by 6.1\% on Prop 30 and 2\% on Prop 37.

\begin{ques}
	\textbf{MRG V.S. ARG}: Is MRG perform better than ARG, or vice verse?
\end{ques}

\begin{myans}
	{
		\emph{
			The two update rules exhibit similar behaviors in terms of accuracy, convergence and running time.}}
\end{myans}

\vspace{0.1cm}
Interestingly, we observe that there is no clear winner between the two update rules. The result demonstrates that our unified algorithm can serve as a framework for comparison between different update rules.


\subsection{Incremental Approach Evaluation}
We justify the advantage of our incremental approaches in terms of guaranteed speed up and decent classification quality compared to the re-computing approach. In the re-computing approach, we apply Algorithm~\ref{alg:framework} to the entire network, on top of old label assignment results. Since the purpose of this group of experiments is to evaluate the incremental framework, not the inference algorithm, we simply choose a representative algorithm MRG in all the following experiments.
\begin{figure}
  \centering
  \includegraphics[width=\columnwidth]{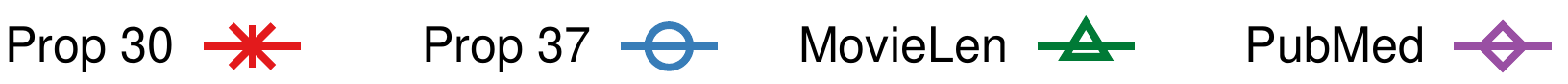}
  \includegraphics[width=0.49\columnwidth]{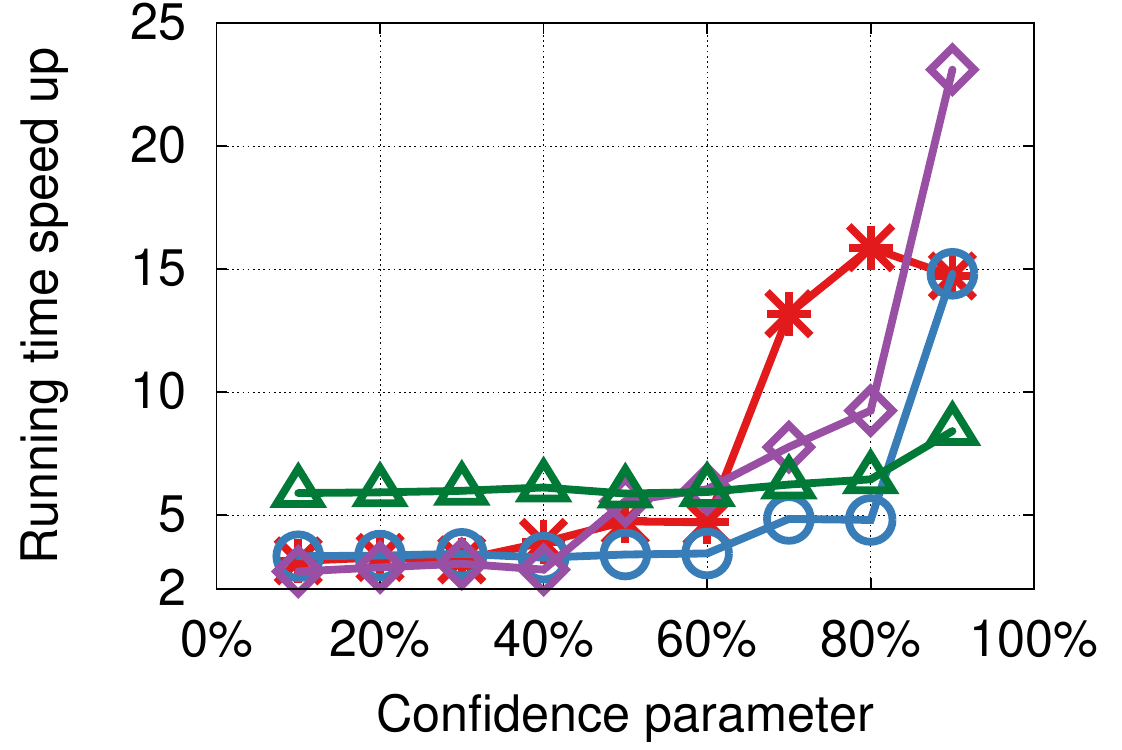}
  \includegraphics[width=0.49\columnwidth]{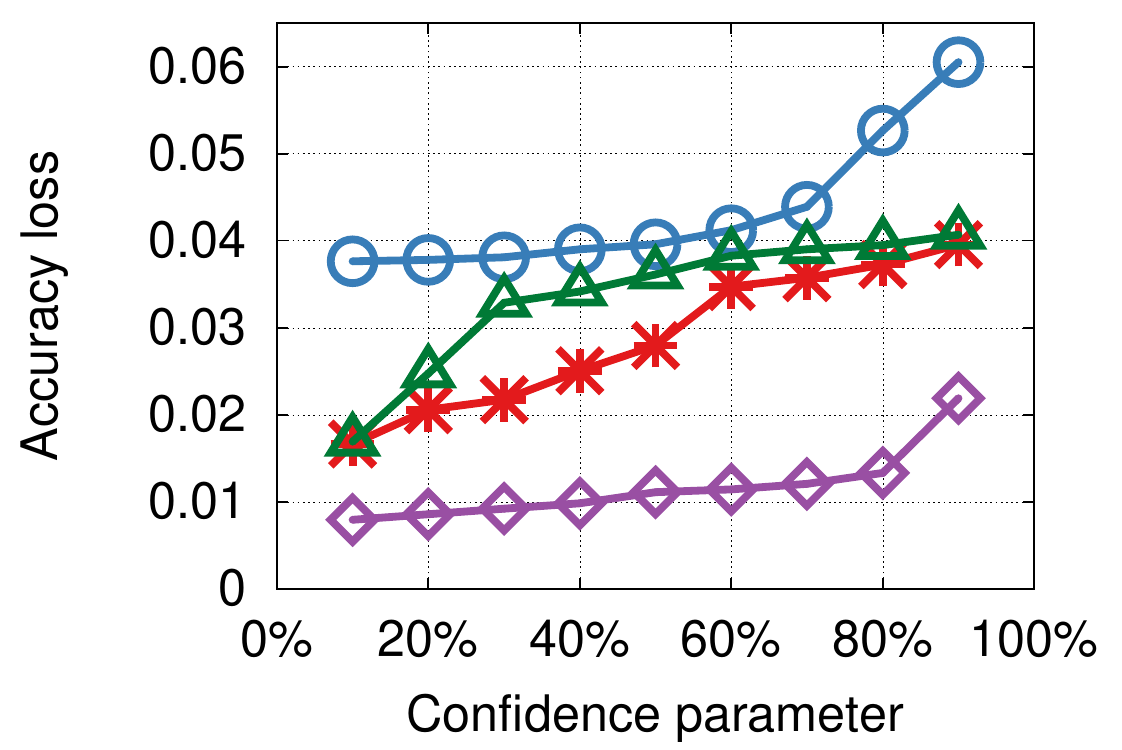}
  \caption{The effect of confidence level parameter $\theta$. Here running time speed up is defined as the running time of re-computing divided by that of incremental approach. The accuracy loss is defined as the accuracy of re-computing minus that of incremental approach.}\label{fig:incremental1}
\end{figure}

\begin{ques}
In terms of both efficiency and quality, what is the effect of confidence parameter $\theta$?
\end{ques}
\begin{myans}
{
\emph{
The running time speed up increases with $\theta$, especially when $\theta>$50\%. On the contrary, the accuracy decreases with $\theta$.}
}
\end{myans}

\vspace{0.1cm}
We first fix the percentage of new data as 10\%, vary the confidence parameter $\theta$ from 0.1 to 0.9, and present the results in Fig.~\ref{fig:incremental1}. Overall the running time speed up is increasing with $\theta$, though the curves have some zigzags due to the ignored effect of $\theta$ on convergence speed. When $\theta$ is greater than 0.5, the running time speed up increases significantly. On the other hand, we do lose information by skipping updates for 100(1-$\theta$)\% of neighbors and thus the accuracy loss is also increased with $\theta$ but much more slowly.

\begin{figure}
  \centering
  \includegraphics[width=\columnwidth]{figure/legenddataset}
  \includegraphics[width=0.49\columnwidth]{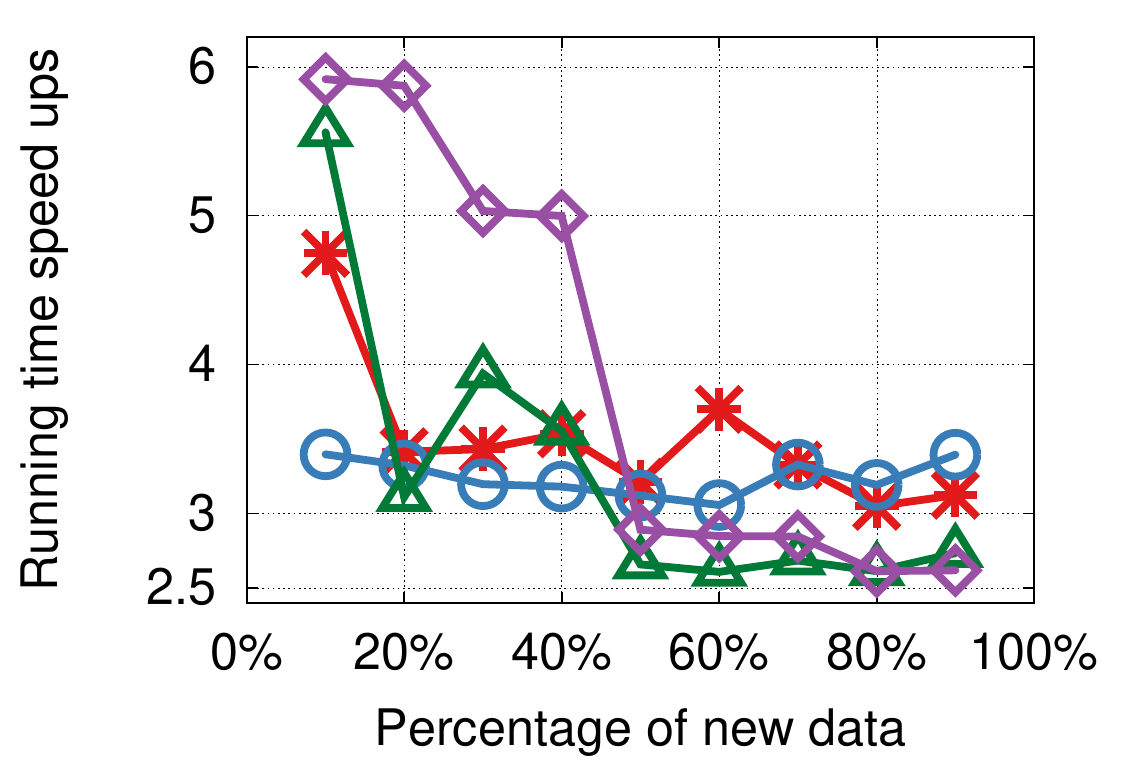}
  \includegraphics[width=0.49\columnwidth]{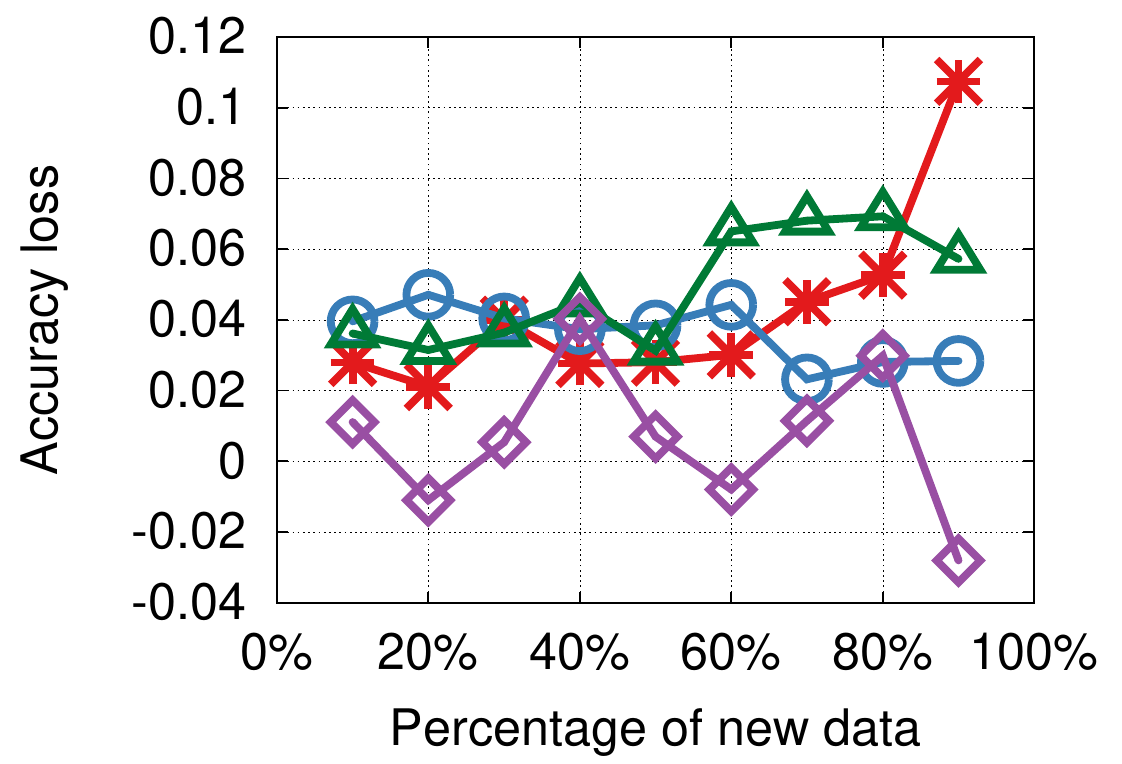}
  \caption{The effect of new data percentage. Note that a\% of new data means that a\% of the entire network is new, in another word, the ratio between old data and new data is $\frac{100-a}{a}$.}\label{fig:incremental2}
\end{figure}

\begin{ques}
What is the effect of new data percentage?
\end{ques}
\begin{myans}
{
\emph{
The efficiency gain obtained by the incremental approach, decreases with the amount of new data. However, it is very interesting to see that applying the incremental approach on a large portion of new data does not necessary lead to a lower accuracy.}
}
\end{myans}

\vspace{0.1cm}
We next fix the confidence parameter $\theta$ as 0.5, vary the percentage of new data from 10\% to 90\%, and report the running time and accuracy in Fig.~\ref{fig:incremental2}. The results clearly indicate that the relative speed up is around 3.5 to 6 when we apply the incremental approach with 10\% of new data. When the percentage of new data is larger than that of old data, the relative speed up decreases to around 2.4 to 3.4. We initially expect that the accuracy loss might also increase when the new data become dominant. However, Fig.~\ref{fig:incremental2} reports that on Prop 37 and PubMed, the accuracy of applying incremental updates on 90\% new data is similar to that on 10\% new data (or even better than). Therefore, it is non-trivial to make a simple decision when and on what condition we should favor the incremental approach purely based on the percentage of new data. We should further examine the similarity between old and new data, as suggested by the utility function defined in Eq.~(\ref{equ:utility}).

\begin{table}[!t]
\centering
\caption{Utility values of the incremental algorithm with 10 different subsets of 10\% new data on Prop 37.}\label{tab:utility}
\small{
\begin{tabular}{|c|c|c|c|}
    \hline
speed ups&accuracy loss&utility (mean)&utility (std)\\
\hline
3.60&-0.060&226.379&78.03\\
\hline
3.71&-0.058&220.828&74.37\\
\hline
2.93&-0.034&218.287&72.15\\
\hline
3.71&-0.033&218.010&77.79\\
\hline
3.43&-0.026&207.193&75.70\\
\hline
2.90&0.000&204.959&72.65\\
\hline
3.08&0.002&201.394&75.85\\
\hline
3.35&0.006&199.478&75.17\\
\hline
3.76&0.012&189.649&73.62\\
\hline
3.76&0.034&186.631&68.56\\
\hline
\end{tabular}
}
\end{table}

\begin{ques}
Is the utility function helpful in making decisions about when we should use the incremental approach?
\end{ques}
\begin{myans}
{
\emph{
The priority of an incremental approach, is correlated with its utility value. A high utility score for an incremental approach suggests a stronger preference to the incremental approach.}
}
\end{myans}

\vspace{0.1cm}
In this set of experiments, we evaluate the goodness of utility functions. We  firstassume that accuracy loss is more concerned and thus we set $u_a=100$ and $u_{s}=60$. We randomly select 10 different subsets of 10\% new data on Prop 37, and then apply both incremental algorithm and recomputing algorithm. For the confidence parameter in the incremental algorithm, we vary it from 0.1 to 0.5. The average running time speed up, the average accuracy loss, the average and standard deviation of utility function are reported in Table~\ref{tab:utility}. Clearly, the utility function is highly correlated with the accuracy loss: a higher utility score leads to less loss in accuracy. Therefore, once we fixed the parameters in the utility function according to application requirements, we are able to safely decide whether we should use the incremental approach based on the utility scores. For example, in Table~\ref{tab:utility}, a straightforward strategy is that if the utility score is greater than 200, we opt to use an incremental approach; otherwise, a re-computing approach is preferred. Moreover, the results also indicate that the percentage of new data is not a good utility measure: We have the same percentage of new data, but accuracy loss, speed up and utility scores vary significantly. We have also changed the setting to $u_a=60$ and $u_s=100$ (presented in Table~\ref{tab:utility2} in Appendix), and the results show that the utility function is highly correlated with the speed up.

\vspace{-3mm}
\section{Conclusions and Future Works}
In this work, we studied the problem of label propagation in $\mathcal{K}$-partite graphs. We proposed a rich label propagation model that supports both heterogeneity and heterophily propagation, by allowing two connected nodes of different types have either similar or opposite labels. We developed a unified label inference framework with two representative update rules. In order to support the dynamic property of real networks, we further presented a fast algorithm, which incrementally assigns labels to new data, or updates old labels according to new feedbacks. Instead of using the percentage of new data, a utility function was further designed to determine when incremental approach is favored.

In the future, we plan to develop an effective adaptive seeding approach, which selects the minimum number of vertices to be labeled but achieves highest guaranteed accuracy. This will greatly reduce the effort and expense of human labeling. 
\section*{Appendix}

%
%

\subsection{Proof of Lemma~\ref{lemma:MAY}}\label{proof:2}
We first consider the general solution to Eq.~(\ref{eq:obj}) without the regularization function. With the observed graph structure, the objective in Eq.~(\ref{eq:obj}) without regularization term can be rewritten as the following dual form:
\begin{equation}\label{eq:objdual}
\small{
\begin{aligned}
&\arg\min_{Y, B}\{\sum_{(u,v)\in E}(G(u,v)-Y(u)^TB_{t(u)t(v)}Y(v))^2\\
&+\sum_{\substack{(u,v)\not \in E,\\
 t(u)\neq t(v)}}(Y(u)^TB_{t(u)t(v)}Y(v))^2\\
&+\beta\sum_{u\in V^L}\|Y(u)-Y^*(u)\|_F^2\}
\end{aligned}
}
\end{equation}
where we separate the heterogenous node pairs into two parts: linked parts and non-linked parts. Note that homogeneous node pairs (pairs of nodes of the same type) are omitted since they are always unlinked in the $\mathcal{K}$-partite graph.

As introduced earlier, we perform vertex-centric update for $Y$. That is, in each iteration, we focus on minimizing the following sub objective:
\begin{equation}
\small{
\begin{aligned}\label{equ:subobj}
J(Y(u))&=\sum_{v\in N(u)}(G(u,v)-Y(u)^TB_{t(u)t(v)}Y(v))^2\\
&+\sum_{v\not\in N(u), t(v)\neq t(u)}(Y(u)^TB_{t(u)t(v)}Y(v))^2\\
&+\mathbf{1}_{V^L}(u)\beta\|Y(u)-Y^*(u)\|_F^2
\end{aligned}
}
\end{equation}
where $1_A(x)$ is the indicator function, which is one if $x\in A$ and zero otherwise.

We introduce the Largrangian multiplier $\Lambda$ for non-negative constraint (i.e., $Y(u)\geq 0$) in Eq.~(\ref{equ:subobj}), which leads to the following Largrangian function $\mathcal{J}(Y(u))$:
\begin{equation*}
\small{
\begin{aligned}
\mathcal{J}(Y(u))&=\sum_{v\in N(u)}(G(u,v)-Y(u)^TB_{t(u)t(v)}Y(v))^2\\
&+\sum_{v\not\in N(u), t(v)\neq t(u)}(Y(u)^TB_{t(u)t(v)}Y(v))^2\\
&+\mathbf{1}_{V^L}(u)\|Y(u)-Y^*(u)\|_F^2+\texttt{tr}(\Lambda_{Y(u)}Y(u)^T)
\end{aligned}
}
\end{equation*}
The next step is to optimize the above terms w.r.t. $Y(u)$. We set $\nabla Y(u)$=0, and obtain:
\begin{equation*}
\small{
\begin{aligned}
\Lambda_{Y(u)}=&-2(\sum_{v\not\in V_{t(u)}} B_{t(u)t(v)}Y(v)Y(v)^TB_{t(u)t(v)}^T+\mathbf{1}_{V^L}(u)I^{k})Y(u)\\
&+2(\mathbf{1}_{V^L}(u)Y^*(u)+\sum_{v\in N(u)}G(u,v)B_{t(u)t(v)}Y(v))
\end{aligned}
}
\end{equation*}

Using the KKT condition $\Lambda_{Y(u)}\circ Y(u)$=0~\cite{kuhn50nonlinear}, where $\circ$ denotes the element-wise multiplicative, we obtain:
\begin{equation*}
\small{
\begin{aligned}
&[-(\sum_{v\not\in V_{t(u)}} B_{t(u)t(v)}Y(v)Y(v)^TB_{t(u)t(v)}^T+\mathbf{1}_{V^L}(u)I^{k})Y(u))\\
&+(\mathbf{1}_{V^L}(u)Y^*(u)+\sum_{v\in N(u)}G(u,v)B_{t(u)t(v)}Y(v))]\circ Y(u)=0\\
\end{aligned}
}
\end{equation*}

Following the updating rules proposed and proved in \cite{DingKDD2006} \cite{GuKDD2009} \cite{ZhuGCL14}, we have:
\begin{equation*}
\small{
\begin{aligned}
&Y(u)=Y(u)\circ\\
&\sqrt{\frac{\sum_{v\in N(u)}G(u,v)B_{t(u)t(v)}Y(v)+\mathbf{1}_{V^L}(u)Y^*(u)}{A_{t(u)}Y(u)+\mathbf{1}_{V^L}(u)Y(u)+\epsilon}}
\end{aligned}
}
\end{equation*}
where $\epsilon>0$ is a very small positive value (e.g., $1^{-9}$), and $A_t$ is defined as follows:
\begin{equation*}
\small{
A_t=\sum_{v\not\in V_{t}}B_{tt(v)}Y(v)Y(v)^TB_{tt(v)}^T
}
\end{equation*}
Note that in the above equation and in all of the following equations, we add a very small positive value $\epsilon$ into denominator to avoid zero division.

This completes the proof.

\subsection{Lipschitz constant for $\nabla J(Y(u))$}\label{proof:constant}
Given a convex function $f$, $\nabla$ f is Lipschitz continuous on Dom $h$ if:
\begin{equation*}
\small{
||\nabla f(x)-\nabla f(y)||_2\leq \mathcal{L}||x-y||_2, \mbox{$\forall$ $x$, $y$ in Dom $h$}
}
\end{equation*}

Therefore, given any two $y_1$ and $y_2$, which denotes two different values for $Y(u)$, we have:
\begin{equation*}
\small{
\begin{aligned}
||\nabla J(y_1)-\nabla J(y_2)||_2=&||2A_t(y_1-y_2)||_2\\
\leq ||2A_t||_2||y_1-y_2||_2
\end{aligned}
}
\end{equation*}

We then have $||\nabla J(y_1)-\nabla J(y_2)||_F\leq 2||A_t||_F||y_1-y_2||_F$, and thus the lipschitz constant for $\nabla J(Y(u))$ is $2||A_t||_F$.

\subsection{Proof of Lemma~\ref{lemma:MAB}}\label{proof:4}
Similar to the proof for Lemma~\ref{lemma:MAY}, we first introduce the Largrangian multiplier $\Lambda$ for non-negative constraint (i.e., $B\geq 0$), which leads to the following Largrangian function $\mathcal{J}(B)$:
\begin{equation}
\small{
\mathcal{J}(B_{tt^{\prime}})=\|G_{tt'}-Y_tB_{tt'}Y_{t'}^T\|^2+\texttt{tr}(\Lambda_{B_{tt'}}B_{tt'}^T)
}
\end{equation}
The next step is to optimize the above terms w.r.t. $B_{tt'}$. We set $\nabla\mathcal{J}(B_{tt'})$=0, and obtain:

\begin{equation*}
\small{
\Lambda_{B_{tt'}}=2Y_{t}^TG_{tt'}Y_{t'}-2Y_{t}^TY_tB_{tt'}Y_{t'}^TY_{t'}
}
\end{equation*}

With the K.K.T. condition~\cite{kuhn50nonlinear}, $\Lambda_{B_{tt'}}\circ B_{tt'}$=0, we have:
\begin{equation*}
\small{
(Y_{t}^TG_{tt'}Y_{t'}-Y_{t}^TY_tB_{tt'}Y_{t'}^TY_{t'})\circ B_{tt'}=0
}
\end{equation*}
Following the updating rules proposed and proved in \cite{DingKDD2006} \cite{GuKDD2009} \cite{ZhuGCL14}, we have:

\begin{equation}
\small{
B_{tt^{\prime}}=B_{tt^{\prime}}\circ\sqrt{\frac{Y_t^TG_{tt^{\prime}}Y_{t^{\prime}}}{Y_t^TY_tB_{tt^{\prime}}Y_{t^{\prime}}^TY_{t^{\prime}}}}
}
\end{equation}
This completes the proof.
\subsection{Lipschitz constant for $\nabla J(B)$}\label{proof:constant2}
Similar to the proof shown in Section~\ref{proof:constant}, given any $B_1$, $B_2$ representing different values of $B_{tt'}$, we have:
\begin{equation*}
\small{
\begin{aligned}
&||\nabla J(B_1)-\nabla J(B_2)||_2\\
=&||2Y_t^TY_tB_1Y_{t'}^TY_{t'}-2Y_t^TY_tB_2Y_{t'}^TY_{t'}||_2\\
\leq &\texttt{tr}(2Y_{t'}^TY_{t'}Y_t^TY_t)||B_1-B_2||_2
\end{aligned}
}
\end{equation*}
 We then have $||\nabla J(B_1)-\nabla J(B_2)||_F\leq 2||Y_{t'}^TY_{t'}Y_t^TY_t||_F||B_1-B_2||_F$, and thus the lipschitz constant is $2||Y_{t'}^TY_{t'}Y_t^TY_t||_F$.

\subsection{Proof of Lemma~\ref{lemma:identical}}\label{proof:identical}
\begin{lemma}\label{lemma:identical}
Updating label assignment $Y$ vertex by vertex using Lemma~\ref{lemma:MAY} is identical to the following traditional multiplicative rule~\cite{ZhuGCL14}:
\begin{equation*}
Y_{t}=Y_t\circ\sqrt{\frac{\sum_{t^{\prime}\neq t}G_{tt^{\prime}}Y_tB_{tt^{\prime}}^T+\beta S_tY_0}{\sum_{t^{\prime}\neq t}Y_tB_{tt^{\prime}}Y_{t^{\prime}}^TY_{t^{\prime}}B_{tt^{\prime}}^T+\beta S_tY_t}}
\end{equation*}
where $S\in R^{n\times n}$ is the label indicator matrix, of which $S_{uu}=1$ if $u\in V^L$ and zero for all the other entries, and $S_t$ is the sub matrix of $S$ for $t$-type vertices.
\end{lemma}
Substituting the multipliers in the preliminary update rule proposed by Zhu et. al.~\cite{ZhuGCL14}, we obtain an optimization algorithm which iterates the following multiplicative update rule for $Y_t$:
\begin{equation*}
\small{
Y_{t}=Y_t\circ\sqrt{\frac{\sum_{t^{\prime}\neq t}G_{tt^{\prime}}Y_{t'}B_{tt^{\prime}}^T+S_tY_0}{\sum_{t^{\prime}\neq t}Y_tB_{tt^{\prime}}Y_{t^{\prime}}^TY_{t^{\prime}}B_{tt^{\prime}}^T+S_tY_t}}
}
\end{equation*}

Note that for each vertex $u$, we have $Y(u)=\texttt{col}(Y^T,u)$, where $\texttt{col} (A, i)$ denotes the specific $i^{\texttt{th}}$-column of a matrix $A$. Therefore, we have $Y(u)$:
\begin{equation*}
\small{
\begin{aligned}
=&Y(u)\circ\sqrt{\frac{\texttt{col}(\sum_{t'\neq t(u)}B_{tt'}Y_{t'}^TG_{t(u)t'}^T, u)+\mathbf{1}_{V^L}(u)Y^*(u)}{\texttt{col}(\sum_{t'\neq t(u)}(B_{tt'}Y_{t'}^TY_{t'}B_{tt'}^T)^TY_{t(u)}^T,u)+\mathbf{1}_{V^L}(u)Y(u)}}\\
=&Y(u)\circ\sqrt{\frac{\sum_{t'\neq t(u')}B_{tt'}Y_{t'}^T\texttt{col}(G_{t(u)t'}^T, u)+\mathbf{1}_{V^L}(u)Y^*(u)}{\sum_{t'\neq t(u)}(B_{tt'}Y_{t'}^TY_{t'}B_{tt'}^T)^T\texttt{col}(Y_{t(u)}^T,u)+\mathbf{1}_{V^L}(u)Y(u)}}\\
=&Y(u)\circ \sqrt{\frac{\sum_{v\in N(u)}G(u,v)B_{t(u)t(v)}Y(v)+\mathbf{1}_{V^L}(u)Y^*(u)}{\sum_{v\not
\in V_{t(u)}}B_{t(u)t(v)}Y(v)Y(v)^TB_{t(u)t(v)}^TY(u)+\mathbf{1}_{V^L}(u)Y(u)}}
\end{aligned}
}
\end{equation*}

Cache the term $\sum_{v\not\in V_t}B_{tt(v)}Y(v)Y(v)^TB_{tt(v)}^T$ as $A_t$, and add $\epsilon$ to the denominator, we have:
\begin{equation*}
\small{
Y(u)=Y(u)\circ\sqrt{\frac{\sum_{v\in N(u)}G(u,v)B_{t(u)t(v)}Y(v)+\mathbf{1}_{V^L}(u)Y^*(u)}{A_{t(u)}Y(u)+\mathbf{1}_{V^L}(u)Y(u)+\epsilon}}
}
\end{equation*}

This is identical to the update rule proposed in Lemma~\ref{lemma:MAY} and thus it completes the proof.

\subsection{Additional tables}
Table~\ref{tab:utility2} reports additional results on utility functions.
\begin{table}[!hbt]
\centering
\caption{Utility values of the incremental algorithm with 10 different subsets of 10\% new data on Prop 37. We fix the confidence level parameter $\theta$ as 0.5 and set $u_a=60$ and $u_s=100$. We report the running time speed up, accuracy loss, and utility function values for each subset of new data.}\label{tab:utility2}
\begin{tabular}{|c|c|c|c|}
    \hline
speed ups&accuracy loss&utility\\
\hline
3.76&0.063&353.152\\
\hline
3.76&0.083&344.492\\
\hline
3.71&0.016&340.527\\
\hline
3.71&0.039&340.095\\
\hline
3.60&0.009&323.221\\
\hline
3.43&0.043&319.736\\
\hline
3.35&0.060&314.174\\
\hline
3.08&0.050&311.186\\
\hline
2.93&0.039&295.853\\
\hline
2.90&0.047&291.144\\
\hline
\end{tabular}
\end{table}
%
%
%
%

%
\section*{Acknowledgment}
We are very grateful to Dr. Kristina Lerman, and Dr. Wolfgang Gatterbauer for their insightful discussions.



\bibliographystyle{abbrv}
\bibliography{BP}

\begin{thebibliography}{10}

\bibitem{abernethy2010graph}
J.~Abernethy, O.~Chapelle, and C.~Castillo.
\newblock Graph regularization methods for web spam detection.
\newblock {\em Machine Learning}, 81(2):207--225, 2010.

\bibitem{Adamic01friendsand}
L.~A. Adamic and E.~Adar.
\newblock Friends and neighbors on the web.
\newblock {\em SOCIAL NETWORKS}, 25:211--230, 2001.

\bibitem{amer2002logical}
S.~Amer-Yahia, M.~Fernandez, R.~Greer, and D.~Srivastava.
\newblock Logical and physical support for heterogeneous data.
\newblock In {\em CIKM Conference}, pages 270--281. ACM, 2002.

\bibitem{Blum:2004:SLU:1015330.1015429}
A.~Blum, J.~Lafferty, M.~R. Rwebangira, and R.~Reddy.
\newblock Semi-supervised learning using randomized mincuts.
\newblock In {\em ICML Conference}, 2004.

\bibitem{CaiPAMI11}
D.~Cai, X.~He, J.~Han, and T.~S. Huang.
\newblock Graph regularized nonnegative matrix factorization for data
  representation.
\newblock {\em {IEEE} Trans. Pattern Anal. Mach. Intell.}, 33(8):1548--1560,
  2011.

\bibitem{Calamai1987}
P.~H. Calamai and J.~J. Mor{\'e}.
\newblock Projected gradient methods for linearly constrained problems.
\newblock {\em Math. Program.}, 39(1):93--116, 1987.

\bibitem{ChakrabartiFCM14}
D.~Chakrabarti, S.~Funiak, J.~Chang, and S.~A. Macskassy.
\newblock Joint inference of multiple label types in large networks.
\newblock In {\em ICML Conference}, pages 874--882, 2014.

\bibitem{DingKDD2006}
C.~Ding, T.~Li, W.~Peng, and H.~Park.
\newblock Orthogonal nonnegative matrix t-factorizations for clustering.
\newblock In {\em SIGKDD Conference}, pages 126--135. ACM, 2006.

\bibitem{Ding:2009:LPK:1726586.1727013}
C.~Ding, T.~Li, and D.~Wang.
\newblock Label propagation on k-partite graphs.
\newblock In {\em ICMLA Conference}, pages 273--278, 2009.

\bibitem{Faloutsos:2014:LGM:2566486.2576889}
C.~Faloutsos.
\newblock Large graph mining: Patterns, cascades, fraud detection, and
  algorithms.
\newblock In {\em WWW Conference}, pages 1--2, 2014.

\bibitem{felzenszwalb2006efficient}
P.~F. Felzenszwalb and D.~P. Huttenlocher.
\newblock Efficient belief propagation for early vision.
\newblock {\em International journal of computer vision}, 70(1):41--54, 2006.

\bibitem{Gama:2014:SCD:2597757.2523813}
J.~a. Gama, I.~\v{Z}liobait\.{e}, A.~Bifet, M.~Pechenizkiy, and A.~Bouchachia.
\newblock A survey on concept drift adaptation.
\newblock {\em ACM Comput. Surv.}, 46(4):44:1--44:37, 2014.

\bibitem{GatterbauerVLDB2015}
W.~Gatterbauer, S.~G\"{u}nnemann, D.~Koutra, and C.~Faloutsos.
\newblock Linearized and single-pass belief propagation.
\newblock {\em Proc. VLDB Endow.}, 8(5):581--592, 2015.

\bibitem{GilpinKDD2013}
S.~Gilpin, T.~Eliassi-Rad, and I.~Davidson.
\newblock Guided learning for role discovery (glrd): Framework, algorithms, and
  applications.
\newblock In {\em SIGKDD Conference}, pages 113--121, 2013.

\bibitem{goldberg2006seeing}
A.~B. Goldberg and X.~Zhu.
\newblock Seeing stars when there aren't many stars: graph-based
  semi-supervised learning for sentiment categorization.
\newblock In {\em Graph Based Methods for Natural Language Processing}, pages
  45--52, 2006.

\bibitem{GuKDD2009}
Q.~Gu and J.~Zhou.
\newblock Co-clustering on manifolds.
\newblock In {\em SIGKDD Conference}, pages 359--368. ACM, 2009.

\bibitem{GuanTLY12}
N.~Guan, D.~Tao, Z.~Luo, and B.~Yuan.
\newblock Nenmf: An optimal gradient method for nonnegative matrix
  factorization.
\newblock {\em IEEE Trans. on Signal Processing}, pages 2882--2898, 2012.

\bibitem{HothesisNMF}
N.-D. Ho.
\newblock {\em NONNEGATIVE MATRIX FACTORIZATION ALGORITHMS AND APPLICATIONS}.
\newblock PhD thesis, 2008.

\bibitem{ihler2005loopy}
A.~T. Ihler, J.~Iii, and A.~S. Willsky.
\newblock Loopy belief propagation: Convergence and effects of message errors.
\newblock In {\em Journal of Machine Learning Research}, pages 905--936, 2005.

\bibitem{JacobWSDM2014}
Y.~Jacob, L.~Denoyer, and P.~Gallinari.
\newblock Learning latent representations of nodes for classifying in
  heterogeneous social networks.
\newblock In {\em WSDM Conference}, pages 373--382, 2014.

\bibitem{Kim:2014:ANM:2582309.2582329}
J.~Kim, Y.~He, and H.~Park.
\newblock Algorithms for nonnegative matrix and tensor factorizations: A
  unified view based on block coordinate descent framework.
\newblock {\em J. of Global Optimization}, 58(2):285--319, 2014.

\bibitem{kim2008competition}
S.-R. Kim and Y.~Sano.
\newblock The competition numbers of complete tripartite graphs.
\newblock {\em Discrete Applied Mathematics}, 156(18):3522--3524, 2008.

\bibitem{kuhn50nonlinear}
H.~W. Kuhn and A.~W. Tucker.
\newblock Nonlinear programming.
\newblock In {\em Proceedings of the 2nd Berkeley Symposium on Mathematical
  Statistics and Probability}, pages 481--492, 1950.

\bibitem{PubMed}
I.~S. Laboratory.
\newblock Computer department of sharif university of technology, pubmed
  dataset.
\newblock http://isl.ce.sharif.edu/pubmed-dataset/.

\bibitem{LeeNIPS2000}
D.~D. Lee and H.~S. Seung.
\newblock Algorithms for non-negative matrix factorization.
\newblock In {\em NIPS Conference}, pages 556--562. MIT Press, 2000.

\bibitem{LinNC2007}
C.-J. Lin.
\newblock Projected gradient methods for nonnegative matrix factorization.
\newblock {\em Neural Comput.}, 19(10):2756--2779, 2007.

\bibitem{Long:2006:ULK:1150402.1150439}
B.~Long, X.~Wu, Z.~M. Zhang, and P.~S. Yu.
\newblock Unsupervised learning on k-partite graphs.
\newblock In {\em SIGKDD Conference}, pages 317--326, 2006.

\bibitem{Nesterov04}
Y.~Nesterov.
\newblock {\em Introductory lectures on convex optimization : a basic course}.
\newblock Kluwer Academic Publ., 2004.

\bibitem{PeiCS15}
Y.~Pei, N.~Chakraborty, and K.~P. Sycara.
\newblock Nonnegative matrix tri-factorization with graph regularization for
  community detection in social networks.
\newblock In {\em IJCAI}, pages 2083--2089. AAAI Press, 2015.

\bibitem{read2007automatic}
I.~Read and S.~Cox.
\newblock Automatic pitch accent prediction for text-to-speech synthesis.
\newblock In {\em Interspeech}, pages 482--485, 2007.

\bibitem{MovieLenData}
S.~Sen, J.~Vig, and J.~Riedl.
\newblock Tagommenders: Connecting users to items through tags.
\newblock In {\em WWW Conference}, pages 671--680, 2009.

\bibitem{ShiLZSY15}
C.~Shi, Y.~Li, J.~Zhang, Y.~Sun, and P.~S. Yu.
\newblock A survey of heterogeneous information network analysis.
\newblock {\em CoRR}, abs/1511.04854, 2015.

\bibitem{SubramanyaJMLR2011}
A.~Subramanya and J.~Bilmes.
\newblock Semi-supervised learning with measure propagation.
\newblock {\em J. Mach. Learn. Res.}, 12:3311--3370, 2011.

\bibitem{talukdar2009new}
P.~P. Talukdar and K.~Crammer.
\newblock New regularized algorithms for transductive learning.
\newblock In {\em Machine Learning and Knowledge Discovery in Databases}, pages
  442--457. Springer, 2009.

\bibitem{Ugander:2013:BLP:2433396.2433461}
J.~Ugander and L.~Backstrom.
\newblock Balanced label propagation for partitioning massive graphs.
\newblock In {\em WSDM Conference}, pages 507--516, 2013.

\bibitem{Vavasis2009}
S.~A. Vavasis.
\newblock On the complexity of nonnegative matrix factorization.
\newblock {\em J. on Optimization}, 20(3):1364--1377, 2009.

\bibitem{Yamaguchi:2015:OSN:2888116.2888151}
Y.~Yamaguchi, C.~Faloutsos, and H.~Kitagawa.
\newblock Omni-prop: Seamless node classification on arbitrary label
  correlation.
\newblock In {\em AAAI Conference}, pages 3122--3128. AAAI Press, 2015.

\bibitem{YangH15}
P.~Yang and J.~He.
\newblock A graph-based hybrid framework for modeling complex heterogeneity.
\newblock In {\em {ICDM} Conference}, pages 1081--1086, 2015.

\bibitem{yedidia2003understanding}
J.~S. Yedidia, W.~T. Freeman, and Y.~Weiss.
\newblock Understanding belief propagation and its generalizations.
\newblock {\em Exploring artificial intelligence in the new millennium}, 2003.

\bibitem{yu2009recommendation}
C.~Yu, L.~V. Lakshmanan, and S.~Amer-Yahia.
\newblock Recommendation diversification using explanations.
\newblock In {\em IEEE 25th International Conference on Data Engineering},
  pages 1299--1302, 2009.

\bibitem{ZhouBLWS03}
D.~Zhou, O.~Bousquet, T.~N. Lal, J.~Weston, and B.~Sch{\"{o}}lkopf.
\newblock Learning with local and global consistency.
\newblock In {\em NIPS Conference}, pages 321--328, 2003.

\bibitem{ZhuGCL14}
L.~Zhu, A.~Galstyan, J.~Cheng, and K.~Lerman.
\newblock Tripartite graph clustering for dynamic sentiment analysis on social
  media.
\newblock In {\em {SIGMOD} Conference}, pages 1531--1542, 2014.

\bibitem{ZhuTKDE16}
L.~Zhu, D.~Guo, J.~Yin, G.~V. Steeg, and A.~Galstyan.
\newblock Scalable temporal latent space inference for link prediction in
  dynamic social networks.
\newblock {\em IEEE Transactions on Knowledge and Data Engineering},
  28(10):2765--2777, Oct 2016.

\bibitem{Zhu03semi-supervisedlearning}
X.~Zhu, Z.~Ghahramani, and J.~Lafferty.
\newblock Semi-supervised learning using gaussian fields and harmonic
  functions.
\newblock In {\em ICML Conference}, pages 912--919, 2003.

\bibitem{vzliobaite2010learning}
I.~{\v{Z}}liobait{\.e}.
\newblock Learning under concept drift: an overview.
\newblock {\em arXiv preprint arXiv:1010.4784}, 2010.

\end{thebibliography}


\end{document}